\pdfoutput=1

\documentclass[11pt]{article}

\usepackage[]{EMNLP2022}

\usepackage{times}
\usepackage{latexsym}

\usepackage[T1]{fontenc}

\usepackage[utf8]{inputenc}

\usepackage{microtype}

\usepackage{inconsolata}

\usepackage{enumitem}
\usepackage{amsmath}
\usepackage{booktabs}
\usepackage{adjustbox}
\usepackage{multirow}

\newcommand{\METRIC}{\textsc{CLIPBERTScore}}
\newcommand{\METAEVAL}{\textsc{MuFaME}}

\title{Evaluating and Improving Factuality in Multimodal Abstractive Summarization}

\author{David Wan \and Mohit Bansal \\
University of North Carolina at Chapel Hill \\
\texttt{\{davidwan,mbansal\}@cs.unc.edu}
}

\begin{document}
\maketitle
\begin{abstract}
Current metrics for evaluating factuality for abstractive document summarization have achieved high correlations with human judgment, but they do not account for the vision modality and thus are not adequate for vision-and-language 
summarization.
We propose \METRIC{}, a simple weighted combination of CLIPScore \cite{hessel-etal-2021-clipscore} and BERTScore \cite{bert-score} to leverage the robustness and strong factuality detection performance between image-summary and document-summary, respectively. 
Next, due to the lack of meta-evaluation benchmarks to evaluate the quality of multimodal factuality metrics, we collect human judgments of factuality with respect to documents and images.
We show that this simple combination of two metrics in the zero-shot setting achieves higher correlations than existing factuality metrics for document summarization, outperforms an existing multimodal summarization metric, and performs competitively with strong multimodal factuality metrics
specifically fine-tuned for the task.
Our thorough analysis demonstrates the robustness and high correlation of \METRIC{} and its components on four factuality metric-evaluation benchmarks. Finally, we demonstrate two practical downstream applications of our \METRIC{} metric: for selecting important images to focus on during training, and as a reward for reinforcement learning to improve factuality of multimodal summary generation w.r.t automatic and human evaluation.\footnote{Our data and code are publicly available at \url{https://github.com/meetdavidwan/faithful-multimodal-summ}}

\end{abstract}

\section{Introduction}

Multimodal abstractive summarization is the task of generating an abridged text that contains the most important information of the source inputs from various modalities.
This challenging task builds upon the success of document summarization, where the input is only text documents. For document summarization, there has been tremendous progress in improving the quality of the summaries with the help of large pre-trained models \cite{lewis-etal-2020-bart, pmlr-v119-zhang20ae, JMLR:v21:20-074}. However, one crucial problem for such models is hallucination, where the model generates contents that are not present or entailed by the document \cite{maynez-etal-2020-faithfulness, falke-etal-2019-ranking}.

\begin{figure}
    \centering
    \includegraphics[width=0.9\columnwidth]{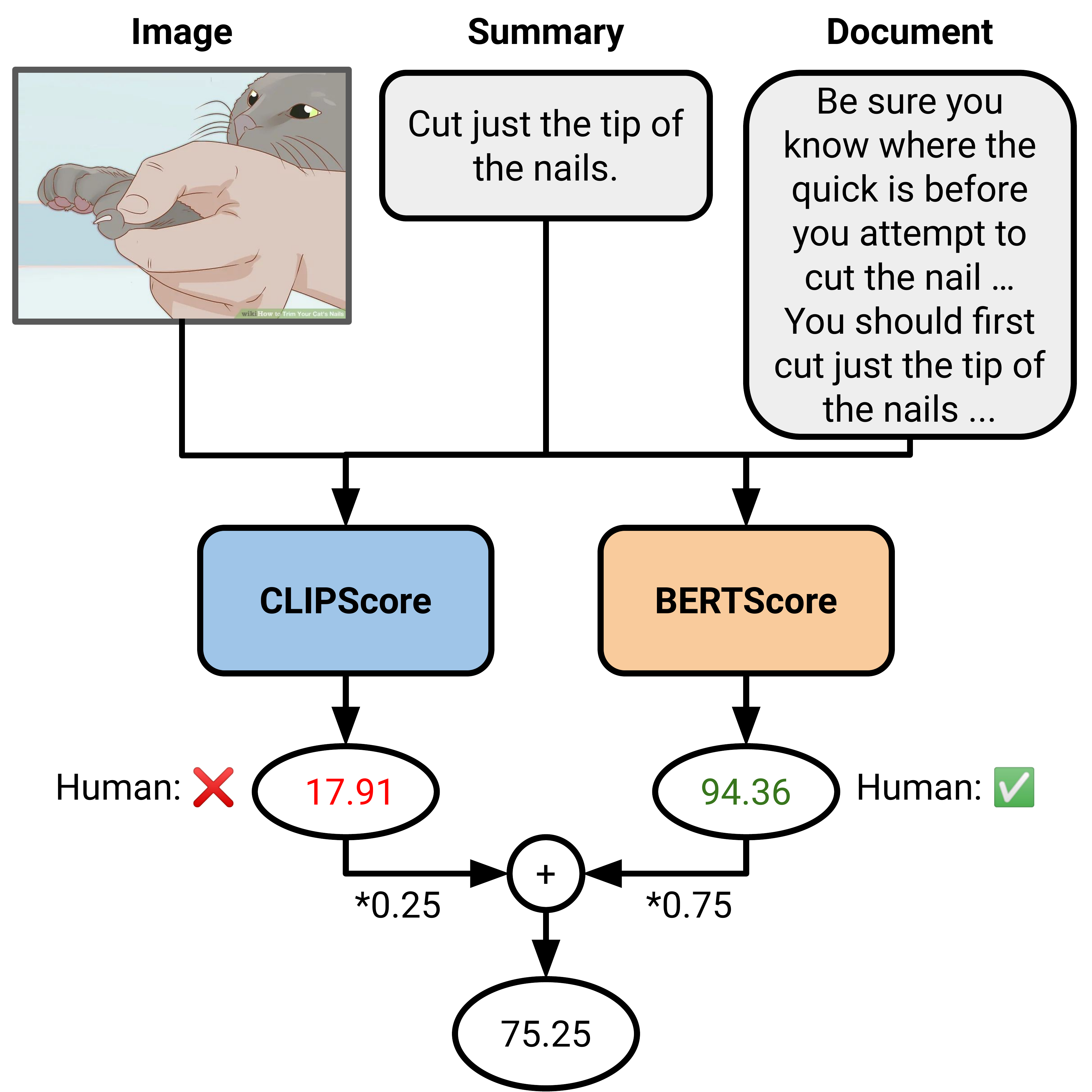}
    \caption{Illustration of the computation of \METRIC{}. CLIP-S and BERT-S computes the image-summary and document-summary factuality score, respectively, and the final score is a weighted combination of the two.
    }
    \label{fig:clipbertscore_figure}
\end{figure}

While there have been significant advancements in developing metrics that correlate highly with the human judgment of factuality \cite{kryscinski-etal-2020-evaluating, durmus-etal-2020-feqa, goyal-durrett-2021-annotating,scialom-etal-2021-questeval}, these metrics only measure factuality between the document and the summary. The lack of judgment between other modalities, such as vision, and the summary makes such metrics not suitable for multimodal settings. We demonstrate this with the example in \autoref{fig:clipbertscore_figure}. The given summary captures less relevant information (cutting the nail) from the document, but it is still considered factual to the document. However, the image shows the main point of the document (finding the place where the nail separates from the quick), making the summary not factual with respect to the image. Current factuality metrics do not account for the image and thus cannot correctly assess factuality for multimodal summaries.

In this work, we introduce a metric that judges factuality of the summary with respect to each input modality. Focusing on the vision-and-language summarization, we propose \METRIC{}, a simple and robust automatic factuality evaluation metric for multimodal summaries that combines two successful metrics: CLIPScore \cite{hessel-etal-2021-clipscore}, which shows strong performance in detecting hallucinations between image and text, and BERTScore \cite{bert-score}, which correlates well with the human judgment of factuality for document summarization \cite{pagnoni-etal-2021-understanding}. 

Next, due to the lack of corpora containing ground-truth human factuality judgments to evaluate multimodal factuality metrics via correlation with human evaluation, we propose a \textbf{Mu}ltimodal \textbf{Fa}ctuality \textbf{M}eta-\textbf{E}valuation (\METAEVAL{}) benchmark by collecting human annotation for four summarization systems and the reference summary on WikiHow,\footnote{\url{https://www.wikihow.com}} a large collection of how-to articles containing rich images relevant to the document. We find that our simple \METRIC{} metric, which combines two off-the-shelf metrics in the zero-shot setting, achieves higher correlation with human judgment over existing text-based factuality metrics, outperforms current multimodal summarization metric MMAE \cite{zhu-etal-2018-msmo}, and performs competitively with a metric trained with a triplet loss specifically for this task.

Next, we perform a detailed analysis of \METRIC{} by evaluating the correlation of the metric and each of its modules on four additional factuality metric-evaluation benchmarks. We first propose the WikiHow Factuality (WikiHowFact) task, derived from the Visual Goal-Step Inference task \cite{yang-etal-2021-visual}. This ranking experiment measures how well the metric can select the correct summary from four choices given the correct document and image.
Since hallucinations are also present for image-captioning task \cite{xiao-wang-2021-hallucination}, we also evaluate the correlations on two captioning tasks focusing on hallucinations, FOIL \cite{shekhar-etal-2017-foil} and BISON \cite{hexiang2018bison}, and one document summarization factuality benchmark FRANK \cite{pagnoni-etal-2021-understanding}. Across all these tasks, we demonstrate the robustness of \METRIC{} and its components, as they achieve the highest correlations compared to strong baselines in each of its respective settings.

Lastly, we present two practical applications for improving the factuality of downstream multimodal summarization models using \METRIC{}: (1) Selecting the most important images as visual guidance \cite{Zhu_Zhou_Zhang_Li_Zong_Li_2020}, and (2) Using the metric as a reward for self-critical sequence training \cite{8099614}. Our results indicate that both applications improve the factuality of the generated summaries across two multimodal summarization datasets, as measured by three factuality metrics and human evaluation.

To summarize, our contributions are:
\begin{enumerate}[leftmargin=*]\setlength{\itemsep}{-1pt}
    \item We propose a simple and robust factuality metric for multimodal summarization based on a combination of CLIPScore and BERTScore.
    \item We create \METAEVAL{}, a meta-evaluation for factuality of multimodal summarization, and the WikiHowFact task to evaluate the quality of multimodal factuality metrics.
    \item We present a detailed study of our metric and its components on various factuality metric-evaluation benchmarks and present strong empirical evidence of its robustness.
    \item We demonstrate two useful downstream applications of our metric to improve the factuality of multimodal abstractive summarization models.
\end{enumerate}

\section{\METRIC{}}\label{sec:model}
\METRIC{} consists of two parts that tackle the image-summary and document-summary factuality judgments, respectively. We show an illustration of the computation in \autoref{fig:clipbertscore_figure}.

\paragraph{Image-Summary.}
We use a variant of CLIPScore \cite{hessel-etal-2021-clipscore} for evaluating the factuality between images and the summary. This metric is based on CLIP \cite{clip}, a cross-modal model trained on 400M image and caption pairs with InfoNCE \cite{infonce} contrastive loss. \citet{hessel-etal-2021-clipscore} finds that using this off-the-shelf model as a metric achieves the highest human correlation on the FOIL \cite{shekhar-etal-2017-foil} benchmark that explores how well the metric can detect hallucinations present in the captions. Thus, it serves as a fitting candidate for factuality evaluation between the image and the summary.

We use CLIP-S, which calculates the cosine similarity between the image embedding $v$ and the text embedding of the summary sentence $t$. To adapt to multimodal summarization, where we have multiple images and multi-sentence summaries,\footnote{The text encoder of CLIP was trained only on single-sentence captions, and the maximum length is set to be 77 tokens. This limits its ability (and that of CLIPScore) to represent multiple sentences.} we take the average of the scores of all image and sentence pairs. Formally, given a list of image embeddings $V$ and summary sentence embeddings $T$ from CLIP's image and text encoder, respectively:
\begin{equation*}
\text{CLIP-S}(V,T) = \frac{1}{|V| |T|} \sum_{i=1}^{|V|} \sum_{j=1}^{|T|} \text{cossim}(v_i,t_j)
\end{equation*}

\paragraph{Document-Summary.} To better detect hallucinations present in the summary with respect to the document, we use the precision variant of BERTScore \cite{bert-score}, which achieves high correlations with human judgments of factuality for document summarization \cite{pagnoni-etal-2021-understanding}. See Section~\ref{sec:frank} for a detailed discussion and comparison against other text-based factuality metrics. Formally, given the contextual embeddings of each token in the document $D$ and summary $S$, it calculates the pairwise cosine similarity between each document and summary token embeddings:
\begin{equation*}
\text{BERT-S} = \frac{1}{|S|} \sum_{s \in S} \max_{d \in D} \text{cossim}(d,s)
\end{equation*}

\paragraph{Full Metric.}
The final score is a combination of the factuality score for image-summary with CLIP-S and that for document-summary with BERT-S: $\text{\METRIC{}} = \alpha \text{CLIP-S} + (1-\alpha) \text{BERT-S}$, where $\alpha$ is a tunable parameter. Please see Section~\ref{sec:weight_combination} for other ways to learn this combination.

\section{Metric Meta-Evaluations}
Next, after defining the multimodal factuality metric \METRIC{}, we want to evaluate the quality of this new metric by checking whether it correlates with human judgments, similar to what has been done for textual factuality metrics \cite{wang-etal-2020-asking,kryscinski-etal-2020-evaluating, durmus-etal-2020-feqa,scialom-etal-2021-questeval}.
As there is no human annotations of factuality for multimodal summarization, we first propose a \textbf{Mu}ltimodal \textbf{Fa}ctuality \textbf{M}eta-\textbf{E}valuation (\METAEVAL{}) benchmark derived from WikiHow to test the correlations of \METRIC{} with human judgments of factuality.

\subsection{\METAEVAL{}}\label{sec:human_eval}
\paragraph{Dataset.} We construct an English multimodal WikiHow summarization dataset \cite{wikihow} for the human evaluation, as this datasets has been extensively studied for document summarization \cite{wikihow, ladhak-etal-2020-wikilingua}, and the images associated with the how-to-articles are relevant to the text. We use a recent WikiHow collection effort by \citet{yang-etal-2021-visual} containing images.\footnote{\url{https://github.com/YueYANG1996/wikiHow-VGSI}. We initially attempted to crawl and align images to the original summarization dataset, but many of the links to the articles are no longer valid or the contents have changed since the original construction.} We generate the step-level multimodal WikiHow dataset by breaking each article into steps and following the construction described in \citet{wikihow}: We consider the first sentence of a step as the summary and the rest of the paragraph as the document, and add the corresponding image. We randomly select 6,000 articles as the validation and test set each, and break each example into steps.\footnote{Although we are primarily interested in the step-level summarization setup for annotation purpose, this creation process also allows future works to experiment with the article-level summarization task by concatenating all the summaries, documents and images of an article.} Statistics of the dataset can be found in \autoref{tab:dataset_statistics} of the Appendix.
For annotation, we randomly sample 50 articles from the test set, and evaluate the generated summaries for all the corresponding steps. Similar to \citet{kryscinski-etal-2020-evaluating}, we split the 50 articles into 10 articles as the validation and 40 as the test set, resulting in 52 and 193 step-level examples for the validation and test set, respectively.

\paragraph{Summarization Systems.} Following \citet{pagnoni-etal-2021-understanding}, we include model summaries from summarization models with varying factuality capabilities. We train four abstractive summarization systems on the multimodal WikiHow dataset, including two document summarization models, T5 \cite{JMLR:v21:20-074} and PEGASUS \cite{pmlr-v119-zhang20ae}, and two multimodal summarization models, CLIP-BART (see \autoref{sec:downstream_applications}), and MOF \cite{zhu-etal-2018-msmo}. Details of the models are provided in Appendix~\ref{sec:human_annotation_models}. We additionally include the reference summaries, resulting in a total of 260 and 965 examples for the validation and test set, respectively.

\paragraph{Annotations.} We conduct the annotations on Amazon Mechanical Turk\footnote{\url{https://www.mturk.com}} (AMT)  platform. For each HIT, we provide the document and the image and ask the workers to read the five summaries. The workers then need to choose whether each summary is faithful to the document and the image separately. An example of the annotation page can be seen in Appendix~\ref{sec:human_annotation_details}. For high-quality annotations, we first conduct a qualification test, where we compare the annotations from the workers against annotations by the authors. Only the workers who have the same annotations on the selected example can perform the actual annotation task. We further select workers from the United States, who have more than 10,000 HITs approved and an approval rate greater than 98\%. We pay 0.18 USD per task to ensure a $>\$12$ hourly rate. Each task consists of three unique workers, and we take the majority class for the document and image factuality judgments, similar to \citet{pagnoni-etal-2021-understanding}. We consider the summary to be faithful only if it is considered faithful to both document and image. We also experiment beyond binary judgment by taking the average over the two factuality judgment to indicate a summary may be partially faithful to one of the source, which is shown in Appendix~\ref{sec:continous_judgment}.

\paragraph{Inter-Annotator Agreement.} We report Fleiss Kappa $\kappa$ \cite{Fleiss1971MeasuringNS} and percentage $p$ of annotators agreement on the majority class, similar to \citet{durmus-etal-2020-feqa}. We obtain $\kappa = 0.50$, with $p=88.5\%$, indicating a moderate agreement \cite{Landis77}.\footnote{For reference, our agreement values are similar to : \citet{pagnoni-etal-2021-understanding} reports $\kappa=0.58, p=91\%$, and \citet{durmus-etal-2020-feqa} reports $p=76.7\%$ on their respective meta-evaluation annotations of XSum and CNN/DM.}

\subsection{Experimental Setup}\label{sec:meta_eval_experiment}

\paragraph{\METRIC{}.} For CLIP-S, we use the \texttt{RN50x64} visual backbone instead of the \texttt{ViT-B/32} version used in the original metric, as the larger backbone shows a higher correlation on factuality benchmarks. For BERT-S, we choose \texttt{RoBERTa-large-mnli} to compute the contextualized embeddings instead of \texttt{RoBERTa-large} for the same reason. We refer readers to Section~\ref{sec:additonal_benchmarks} for more details. We  use the validation set of \METAEVAL{} to tune $\alpha$, where we find that $\alpha=0.25$ achieves the best correlations on the combined judgment. We use this parameter for all  experiments (See Section~\ref{sec:weight_combination} for other ways to learn this combination).
 
\paragraph{Baseline Metrics.} Having separate judgments for document-summary, image-summary, and multimodal settings allows us to evaluate the metrics' performance with different modality combinations.
For document-summary, we compare against existing factuality metrics, including FactCC \cite{kryscinski-etal-2020-evaluating}, DAE \cite{goyal-durrett-2021-annotating}, QuestEval \cite{scialom-etal-2021-questeval}, and the original BERTScore \cite{bert-score}. We also measure the performance of the text-matching component of CLIPScore, which we refer to as CLIPText-S. 
For image-summary evaluation, we compare our CLIP-S against Triplet Network, as described in \citet{yang-etal-2021-visual}. We train this metric on the multimodal WikiHow dataset, allowing comparisons of correlations between CLIP-S in the zero-shot setting and a fine-tuned metric for this task.
For multimodal factuality metrics, we experiment with several weighted combinations of document-summary and image-summary metrics by tuning the weights on the validation set, including combinations of DAE with CLIP-S, Triplet Network with BERT-S, and RefCLIP-S.
We also compare to MMAE \cite{zhu-etal-2018-msmo} developed for evaluating the summary quality of multimodal summarization. As the metric is originally designed for a different dataset, we similarly use the multimodal WikiHow to train its image-summary component IM\textsubscript{Max}. We refer the readers to Appendix~\ref{sec:metrics_details} for details of the metrics.

\begin{table}[!t]
    \centering
    \small
    \adjustbox{max width=\columnwidth}{
    \begin{tabular}{c c c c c}
    \toprule
    Metric & Document & Image & Combined \\
    \midrule
    FactCC & 0.01 & - & 0.00 \\
    DAE & 0.50 & - & 0.38 \\
    QuestEval & 0.41 & - & 0.32 \\
    CLIPText-S & 0.19 & - & 0.14 \\
    BERTScore & 0.54 & - & 0.40 \\
    BERT-S & \textbf{0.58} & - & \textbf{0.43} \\
    \midrule
    CLIP-S & - & 0.22 & 0.21 \\
    IM\textsubscript{Max} &  - & 0.10 & 0.07 \\
    Triplet Net & - & \textbf{0.25} & \textbf{0.25} \\
    \midrule
    MMAE & 0.21 & 0.26 & 0.22 \\
    RefCLIP-S & 0.20 & 0.26 &  0.25 \\
    CLIP-S + DAE & 0.53 & 0.33 & 0.41 \\
    Triplet Net + BERT-S & 0.58 & \textbf{0.44} & \textbf{0.47} \\
    \METRIC{} & \textbf{0.59} & 0.42 &  \textbf{0.47} \\
    \bottomrule
    \end{tabular}
    }
    \caption{
    Pearson correlation coefficients between automatic metrics and human judgments of factuality with respect to the document, image, and combined. The top section corresponds to factuality metrics for document summarization, the middle section corresponds to image-summary factuality metrics, and the bottom section corresponds to multimodal metrics.}
    \label{tab:human_eval_result}
\end{table}

\subsection{Meta-Evaluation Results}\label{sec:human_eval_result}
\autoref{tab:human_eval_result} shows the Pearson correlation of the automatic metrics. We first note that the combined judgments require taking both modalities into consideration. Metrics that only consider the document correlate less with the combined judgment than with the document-only judgment, indicating the importance of capturing the vision modality for evaluating factuality for multimodal summarization. Multimodal factuality metrics, on the other hand, show positive transfers, as they correlate higher on all three settings than its components.

Next, for the document-summary factuality judgments, BERT-S achieves the highest correlation, outperforming DAE by 8 points and the original BERTScore by 4 points.
Compared to MMAE, which is developed for evaluating the quality of multimodal summarization, \METRIC{} significantly outperforms on all three categories, showing the importance of targeting the factuality aspect.
While Triplet-Net achieves better correlations on image, \METRIC{} actually outperforms the fine-tuned variants for the document case and provides the same correlations on the combined case.
We thus stress the simplicity of \METRIC{} of only requiring the use of two off-the-shelf metrics in the zero-shot setting without the need for extra training to compare competitively with fine-tuned method.

\begin{table}[!t]
    \centering
    \small
    \adjustbox{max width=\columnwidth}{
    \begin{tabular}{c c c c c}
    \toprule
    Metric & Document & Image & Combined \\
    \midrule
    \METRIC{} & 0.59 & \textbf{0.42} & \textbf{0.47} \\
    \METRIC{}\textsubscript{logis} & 0.54 & 0.38 & 0.41 \\
    \METRIC{}\textsubscript{MLP} & \textbf{0.60} & \textbf{0.42} & \textbf{0.47} \\ 
    \bottomrule
    \end{tabular}
    }
    \caption{Meta-evaluation results with different combination methods.}
    \label{tab:weight_combination}
\end{table}

\subsection{Comparison of Combination Strategies}\label{sec:weight_combination}
While \METRIC{} uses $\alpha$ to decide the weights for CLIP-S and BERT-S, we also explore using logistic regression (logis) and multi-layer perceptron (MLP) to output a final score given the two modules, following \citet{zhu-etal-2018-msmo}.\footnote{We also attempted to train a single multimodal evaluation model takes as input both the visual and text features from the image and summary to output the factuality judgment using the InfoNCE \cite{infonce} contrastive loss, but this also performs worse than \METRIC{}.}
Similar to the $\alpha$ parameter, we tune the two methods by fitting the metric on the combined human judgment scores and selecting the parameters that would achieve the highest Pearson correlation on the development set of \METAEVAL{} meta-evaluation dataset.\footnote{We use the re-scaled BERT-S scores for the two methods, as tuning with the original BERT-S achieves low correlations}
The result is presented in \autoref{tab:weight_combination}. While logistic regression performs the worst, using MLP for combining the two modules provides similar correlations as \METRIC{} that uses the $\alpha$ parameter. Specifically, MLP provides a point increase in correlations with respect to the document but provides the same correlations on the combined judgment. The weight combination strategies can be chosen based on preference, but we advocate for the simplicity with the $\alpha$ parameter.

\section{Additional Factuality Metric-Evaluation Benchmarks}\label{sec:additonal_benchmarks}
We evaluate \METRIC{} and its components on additional factuality metric-evaluation benchmarks, focusing on how robust the metrics performs across a variety of tasks and domains.

\begin{figure}
    \centering
    \includegraphics[width=0.9\columnwidth]{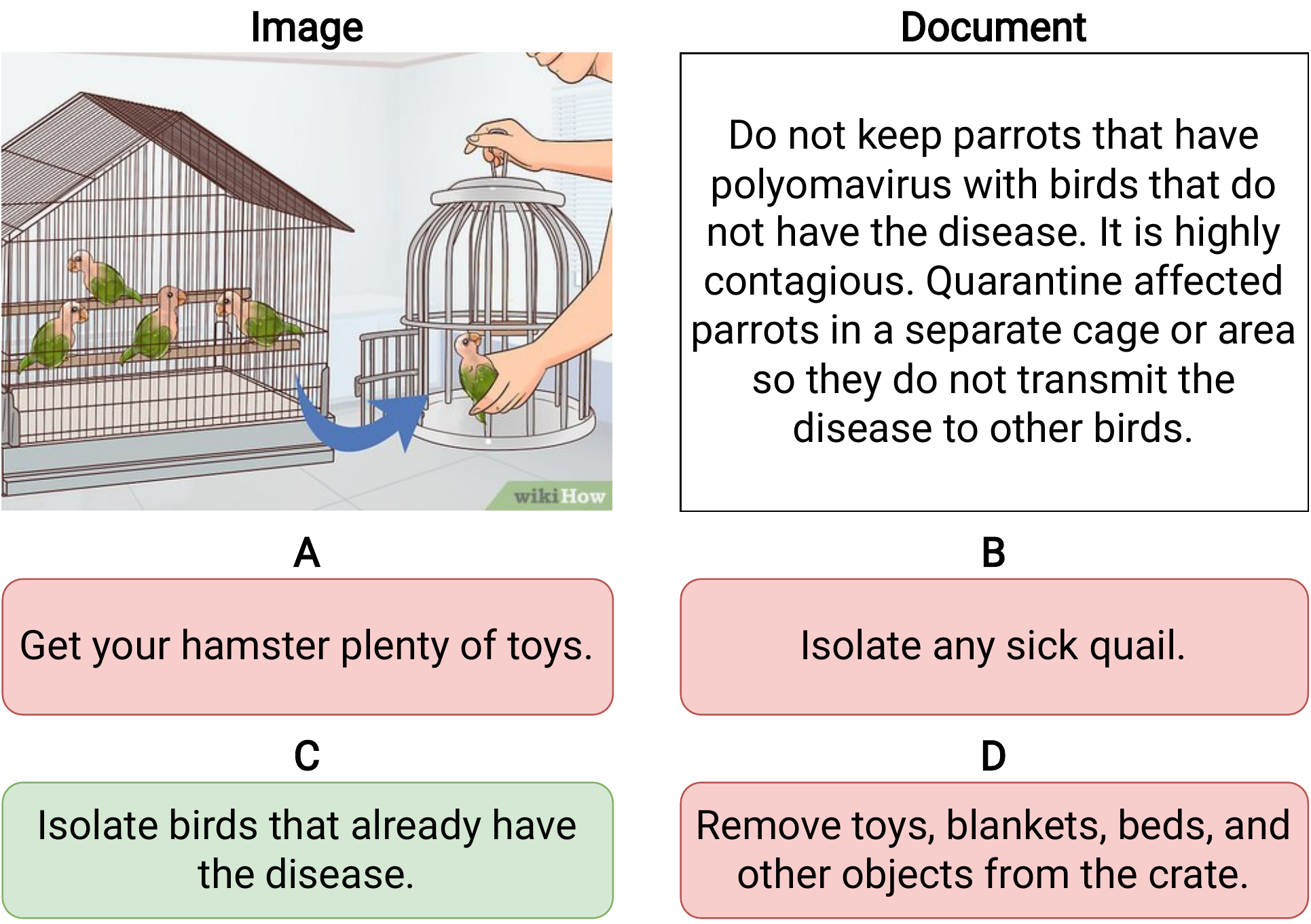}
    \caption{An example of the WikiHowFact task. Given the image and document, the metric needs to select the correct summary C. The task can be split into image-summary and document-summary evaluation by only providing the respective input.}
    \label{fig:wikihow_fact_example}
\end{figure}

\subsection{WikiHow Factuality}\label{sec:wikihow_fact}

We propose the WikiHow Factuality (WikiHowFact) task that evaluates how well the metric can choose the correct summaries over incorrect ones.
We derive this task from WikiHow VGSI \cite{yang-etal-2021-visual} to evaluate the text-image matching performance as a ranking experiment, which has been explored for factuality metric evaluation \cite{falke-etal-2019-ranking}. An example can be seen in \autoref{fig:wikihow_fact_example}. Each example uses the correct document and image as the prompt and includes four choices consisting of the correct summary and three negative summaries generated by \textit{random}, \textit{category}, and \textit{similarity} sampling strategies described in \citet{yang-etal-2021-visual}. We note that this setup is actually a more challenging task than the original VGSI task. See Appendix~\ref{sec:wikihowfact_appendix} for more details.
Similar to the meta-evaluation in Section~\ref{sec:meta_eval_experiment}, we consider the document, image and combined settings depending on the choice of the prompt, and evaluate using the the same sets of metrics. We further compare CLIP-S to that using the smaller \texttt{ViT-B/32} visual backbone. We compute the ranking accuracy of assigning a higher score to the correct summary. See Appendix~\ref{sec:wikihowfact_appendix} for more details.

\begin{table}[!t]
    \centering
    \adjustbox{max width=\columnwidth}{
    \begin{tabular}{c c | c c c}
    \toprule
    Prompt & Metric & Random & Category & Similarity \\
    \midrule
    \multirow{5}{*}{Document} & FactCC & 41.66 & 38.54 & 37.28 \\
    & DAE & 69.51 & 69.54 & 68.37 \\
    & CLIPText-S & \textbf{91.12} & \textbf{87.38} & \textbf{82.81} \\
    & BERTScore & 82.46 & 80.10 & 77.07 \\
    & BERT-S & 86.24 & 84.97 & 81.97 \\
    \midrule
     \multirow{3}{*}{Image} & CLIP-S ViT-B/32 & 78.13 & 69.47 & 53.92 \\
    & Triplet Net & \textbf{82.21} & \textbf{75.23} & \textbf{65.18} \\
    & CLIP-S & 81.14 & 74.47 & 60.24 \\
    \midrule
    \multirow{3}{*}{Combined} & RefCLIP-S & 93.04 & 88.05 & 78.66 \\
    & Triplet Net + BERT-S & 94.27 & 91.99 & \textbf{87.03} \\
    & CLIPBERTScore & \textbf{95.46} & \textbf{92.56} &	85.12 \\
    \bottomrule
    \end{tabular}
    }
    \caption{WikhowFact ranking accuracy given different input modalities. \METRIC{} shows the largtest positive transfers when combined, outperforming RefCLIP-S on all settings and Triplet Net + Bert-S on random and category settings.
    }
    \label{tab:wikihow_fact}
\end{table}

\paragraph{Results.} We present the WikiHowFact result in \autoref{tab:wikihow_fact}. First, for the image-summary setting, we observe the power of larger visual backbone at improving factuality, as CLIP-S achieves a 3, 5, and 6.3 point increase compared to CLIP-S ViT-B/32 for the random, category, and similarity split, respectively. For document-summary setting, CLIPText-S and BERT-S achieve high accuracy across the sampling strategies. Interestingly, CLIPText-S performs better than BERT-S, but this does not apply to the multimodal case: \METRIC{} actually outperforms RefCLIP-S, showing the better positive transfer between CLIP-S and BERT-S.
Similar to the meta-evaluation results, the Triplet Network outperforms CLIP-S for the image-summary setting, but the difference on random and category splits is only around 1 point. \METRIC{} still outperforms Triplet Network + BERT-S on the random and category splits, indicating the strong performance of combining the two metrics for evaluating factuality.

\begin{table}[!t]
    \centering
    \small
    \begin{tabular}{l c c c c}
    \toprule
    Metric & no-ref & 1-ref & 4-ref \\
    \midrule
    length* & - & 50.2 & 50.2 \\
    ROUGE-L* & - & 71.7 & 79.3 \\
    CLIPText-S ViT-B/32 & - & 87.1 & 90.6 \\
    CLIPText-S RN50x64 & - & \textbf{87.8} & 92.0 \\
    BERT-S & - & \textbf{87.8} & \textbf{93.6} \\
    \midrule
    Triplet Net + BERT-S & 60.5 & 85.0 & 90.7 \\
    RefCLIP-S ViT-B/32 & 86.8 & 91.7 & 92.6 \\
    RefCLIP-S RN50x64 & \textbf{90.1} & 92.0 & 93.8 \\
    CLIPBERTScore & \textbf{90.1} & \textbf{92.7} & \textbf{95.0} \\
    \bottomrule
    \end{tabular}
    \caption{FOIL accuracy. * indicates results taken from \citet{hessel-etal-2021-clipscore}. Top section represents correlations of factuality metrics for the document-summary setting, while the bottom section show that for the image-summary setting (no-ref) and multimodal setting.}
    \label{tab:foil}
\end{table}

\subsection{Hallucination in Image Captioning} \label{sec:foil}
The FOIL \cite{shekhar-etal-2017-foil} dataset measures how well the metric can differentiate correct MSCOCO captions from hallucinated ones generated by adversarially swapping out noun phrases. We follow \citet{hessel-etal-2021-clipscore} and evaluate metrics on the paired setting. We compute the ranking accuracy by giving only the image (no-ref), and with 1 or 4 additional reference captions.
We compare \METRIC{} and its components with CLIPScore variants using the \texttt{ViT-B/32} backbone. We refer the readers to Appendix~\ref{sec:foil_appendix} for more details and results on all visual backbones.
We present the results in \autoref{tab:foil}. BERT-S achieves the highest accuracy in terms of the text-matching performance. Especially when more text (4 references) is provided, it outperforms CLIPText-S by 1.6 points. For image-text matching, we observe similar improvement with larger visual backbones. \METRIC{} showcases its strength at positive transfer of its two components: we observe improvement over RefCLIP-S RN50x64 of 0.7 points for 1-ref and 1.2 points for 4-ref.

\subsection{Fine-grained Visual Grounding}
BISON \cite{hexiang2018bison} measures the ability of the metric to select the correct MSCOCO image from a semantically similar image, requiring more fine-grained visual grounding to achieve high accuracy. We compare the image-summary metrics, and refer the readers to Appendix~\ref{sec:bison_appendix} for results on all CLIP-S variants. \autoref{tab:bison} shows that CLIP-S actually outperforms the fully fine-tuned SOTA metric, SCAN t2i \cite{lee2018stacked}, indicating its robustness and strong text-image detection performance in the zero-shot setting. Triplet Network on the other hand is not robust to this task, achieving much lower accuracy than all other metrics.

\subsection{Document Summarization Factuality Evaluation}\label{sec:frank}
We compare how BERT-S and CLIPText-S correlate on FRANK, a factuality benchmark evaluation for document abstractive summarization containing 2,250 annotations for generated summaries on XSum \cite{narayan-etal-2018-dont} and CNN/DM \cite{NIPS2015_afdec700}. We report Pearson and Spearman correlations, using the official evaluation script.\footnote{\url{https://github.com/artidoro/frank}}
The result is shown in \autoref{tab:frank} in the Appendix. CLIPText-S does not perform well for detecting faithful summaries for summarization, as Pearson and Spearman coefficients are around 0.10 for all datasets and 0.05 for XSum Spearman. In contrast, BERT-S that uses RoBERTa \cite{roberta} model finetuned on the MNLI \cite{williams-etal-2018-broad} correlates higher than the original BERTScore on Pearson and Spearman across both datasets. It is thus useful to treat factuality as an entailment problem and use the appropriate model.

\begin{table}[!t]
    \centering
    \small
    \begin{tabular}{l c}
    \toprule
    Metric & Acc \\
    \midrule
    SCAN t2i* & 85.89 \\
    Triplet Net & 63.16 \\
    CLIP-S ViT-B/32 & 83.85 \\
    CLIP-S & \textbf{86.03} \\
    \bottomrule
    \end{tabular}
    \caption{BISON accuracy. * indicates result copied from \citet{hexiang2018bison}.}
    \label{tab:bison}
\end{table}

\begin{table*}[!t]
    \centering
    \adjustbox{max width=\textwidth}{
    \begin{tabular}{l c c c c c c c c c}
    \toprule
    Model & R1 & R2 & RL & IP & BERTScore & FactCC & DAE $\downarrow$ & QuestEval \\
     \midrule
     CLIP-BART ROUGE guidance & \textbf{43.66} & \textbf{20.79} & \textbf{30.42} & 74.25 & 94.21 & 87.60 & 6.31 & 58.79 \\
     CLIP-BART \METRIC{} guidance & 43.52 & 20.67 & 30.27 & \textbf{74.87} & \textbf{94.29} & \textbf{88.47} & \textbf{5.71} & \textbf{58.92}  \\
    \bottomrule
    \end{tabular}
    }
    \caption{ MSMO result with different guidance strategies. DAE: lower is better ($\downarrow$). For reference, the SOTA model UniMS \cite{unims} achieves 42.94 for R1, 20.50 for R2, and 69.38 for image precision (IP).  \METRIC{} as a guidance improves IP and all factuality metrics with a minor decrease in ROUGE.}
    \label{tab:msmo_guidance}
\end{table*}

\section{Downstream Applications}\label{sec:downstream_applications}
Finally, we present two useful downstream applications for improving factuality of multimodal summarization models: first by using the metric as a reference image selection to guide the model in attending important images, and second by using it as a reward for self-critical sequence training.
For both applications, we train strong baseline models by adapting CLIP-BART \cite{sung2022vladapter} for multimodal summarization. Specifically, we extract visual features with CLIP and use a projection layer to transform the dimension of the visual representation to the correct dimension of BART \cite{lewis-etal-2020-bart}. Then, the projected features are concatenated with the text features from the original encoder as the joint input representation for BART. See Appendix~\ref{sec:rl_experiment} for more details.

\subsection{Multimodal Visual Guidance}\label{sec:visual_guidance}
One of the well-known tasks is multimodal summarization with multimodal output \cite[MSMO]{Zhu_Zhou_Zhang_Li_Zong_Li_2020}, which incorporates the associated images with the CNN/DM articles. The authors shows that previous models suffer from modality bias, as the cross-entropy loss is only based on the text modality. To help the model also attend to the vison modality, the authors propose to create visual references by ranking and selecting the most relevant input images. While the authors show improved performance by ranking the images by the ROUGE score between the corresponding caption and the reference summary, such reference does not explicitly guide the model to generate summaries faithful with respect to the images.
We thus propose to use \METRIC{} to select reference images.
To incorporate the visual guidance into the training, we add a guidance loss by minimizing the cross-entropy loss, where we consider the selected images by the reference as correct, and the remaining images as incorrect. We use each image's hidden representation from the encoder to produce a binary prediction using a linear layer.

We compare against the model using ROUGE as the visual guidance. Following \citet{zhu-etal-2018-msmo}, we report the performance of the models on ROUGE, and image precision (IP) of the model's recommended images and human-annotated relevant images. We additionally evaluate factuality using BertScore, FactCC, DAE, and QuestEval.
The result is shown in \autoref{tab:msmo_guidance}. We observe a correlation between the guidance metric and the metric score, as the model with ROUGE guidance achieves higher ROUGE scores, and the model with \METRIC{} guidance improves all factuality metrics and IP. Though the gain is relatively
small, the improvement on factuality metrics is greater than the negligible drop in ROUGE.

\begin{table*}[t!]
    \centering
    \small
    \begin{tabular}{l l c c c c c c c c c c}
    \toprule
     Dataset & Model & R1 & R2 & RL & IP & BERTScore & FactCC & DAE $\downarrow$ & QuestEval \\
    \midrule
    \multirow{2}{*}{MMSS} &  Base & 55.95 & 32.18 & 51.81 & - & 93.59 & 66.75 & 12.20 & 54.37 &  \\
    & Base + Reward & \textbf{55.99} & \textbf{32.60} & \textbf{52.05} & - & \textbf{94.79} & \textbf{71.60} & \textbf{$\phantom{0}$6.60} & \textbf{58.59} & \textbf{} \\
    \midrule
    \multirow{2}{*}{MSMO} &  Base & 43.52 & 20.67 & \textbf{30.27} & \textbf{74.87} & 94.29 & 88.47 & $\phantom{0}$5.71 & 58.92 \\
    & Base + Reward & \textbf{43.87} & \textbf{20.92} & 30.04 & 74.31 & \textbf{94.87} & \textbf{94.96} & \textbf{$\phantom{0}$0.61} & \textbf{60.48} \\
    \bottomrule
    \end{tabular}
    \caption{SCST result on MMSS and MSMO. DAE: lower is better ($\downarrow$). We train a CLIP-BART model as the base model for MMSS, and use CLIP-BART \METRIC{} guidance as the base model for MSMO. We observe consistent improvement on all metrics with SCST over the base model on MMSS, and consistent improvement on all factuality metrics on MSMO. For reference, the SOTA model on MMMS by \citet{li-etal-2020-multimodal} achieves 48.19/25.64/45.27 for ROUGE.}
    \label{tab:rl_result}
\end{table*}

\begin{table}[!t]
    \centering
    \small
    \begin{tabular}{l c c c}
    \toprule
     Model & Document & Image & Comb \\
     \midrule
     Base & 88.67$\phantom{**}$ & 52.67$\phantom{*}$ & 70.67$\phantom{**}$ \\
     Base + Reward & \textbf{95.00**} & \textbf{55.00*} & \textbf{75.00**} \\
     \bottomrule
    \end{tabular}
    \caption{Human evaluation results on MMSS. Model with SCST training are statistically signficiantly more factual with respect to document, image, and both. * indicates $p< 0.05$ and ** indicates $p<0.01$.}
    \label{tab:mmss_human_eval}
\end{table}

\subsection{Self-Critical Sequence Training with \METRIC{} Reward}
A more generalized application to improve factuality is to use \METRIC{} as a reward for the self-critical sequence training \cite[SCST]{8099614}, which optimizes the model using the REINFORCE algorithm \cite{10.1007/BF00992696}. Formally, given document d, images v, and the summary y, the self critical loss is defined as:
\begin{equation*}
 \resizebox{\columnwidth}{!}{$L_{SCST} = -(r(y^s) - r(\hat{y})) \sum_{t=1}^{N} \log{P (y_t^s|y_1^s,...,y_{t-1}^s, d,v)} $
 }
\end{equation*}
where $r(\cdot)$ is a reward function, $y^s$ is the sampled summary, and $\hat{y}$ is the summary obtained by greedy decoding. We follow previous works \cite{pasunuru-bansal-2018-multi,li-etal-2019-deep,parnell-etal-2021-rewardsofsum} and train on the combined loss of cross-entropy $L_{XE}$ and the self-critical loss: $L = \alpha L_{RL} + (1-\alpha)L_{XE}$, where we set $\alpha = 0.998$.

We use \METRIC{} and ROUGE-2 as the rewards, so as to improve factually while maintaining informativeness. Following \citet{pasunuru-bansal-2018-multi}, we alternate the rewards for each step during the training. We upweight \METRIC{} by 2x (tuned on the validation set).
We experiment on MSMO, and the multimodal sentence summarization \cite[MMSS]{ijcai2018-577} task, which combines the Gigaword corpus \cite{graff2003english,rush-etal-2015-neural} with crawled images.\footnote{Since there is only one image associated with each example for MMSS, we do not add visual guidance for models trained on this dataset.}
As the base models, we use the CLIP-BART + \METRIC{} model trained in Section~\ref{sec:visual_guidance} for MSMO, and we similarly train a CLIP-BART model for the MMSS. We then use the fine-tuned models and train with SCST. Details of the training details can be found in Appendix~\ref{sec:rl_experiment}. We report the same metrics for MMSS except for IP, since the task does not contain gold image labels.

The result is shown in \autoref{tab:rl_result}. We see consistent improvement over all metrics with SCST for MMSS. Specifically, we observe a 5-point improvement for FactCC and DAE, and a 4-point increase for QuestEval while maintaing similar ROUGE score. We observe a similar trend for training with SCST on MSMO dataset, where SCST training improves FactCC, DAE and QuestEval by 8 points, 5 points, and 1.5 points, respectively.

To evaluate the factuality of the summaries generated by models trained with SCST against that by the base model, we conduct a human evaluation on a randomly sampled 100 articles from the MMSS test set. We perform the same AMT experiment as described in Section~\ref{sec:human_eval}. We ensure the same $> \$12$ hourly rate and pay 0.1 USD per HIT. For each summary, we aggregate the 3 annotator scores for the document, image, and combined judgments. The final factuality score is the average across the 100 examples. The result is shown in \autoref{tab:mmss_human_eval}. The model with SCST training achieves a statistically significantly better factuality score with respect to the document ($p=0.002$), image ($p=0.041$), and especially to the combined case ($p<0.001$) using bootstrap test \cite{EfroTibs93}. This aligns with the factuality improvement we observe with the automatic factuality scores in \autoref{tab:rl_result}.

\section{Related Work}

\paragraph{Multimodal Summarization.} The task of multimodal summarization takes additional inputs from multiple modalities apart from the input text document, including images \cite{ijcai2018-577, Zhu_Zhou_Zhang_Li_Zong_Li_2020,Li_Yuan_Xu_Wu_He_Zhou_2020} and videos \cite{li-etal-2020-vmsmo, palaskar-etal-2019-multimodal}. To incorporate multiple modalities, many works have developed models with multimodal attention \cite{Zhu_Zhou_Zhang_Li_Zong_Li_2020}. When multiple images are present, the rich information present in the images may distract and thus hurt the model's performance. To this end, approaches such as selective gating \cite{li-etal-2020-multimodal}, visual guidance \cite{Zhu_Zhou_Zhang_Li_Zong_Li_2020}, and knowledge distillation \citet{unims} have been proposed. While these methods have demonstrated improvement in ROUGE, to the best of our knowledge, factuality for such tasks has not been studied. We aim to provide an evaluation benchmark for evaluating factuality, and demonstrate methods to improve factuality for the multimodal summarization task.

\paragraph{Faithfulness and Factuality Metrics.} Many metrics have been proposed to evaluate the factuality of generated summaries. The metrics can be roughly categorized into entailment-based and question generation and question answering (QGQA) metrics. Entailment-based metrics \cite{kryscinski-etal-2020-evaluating, goyal-durrett-2021-annotating} train metrics to predict entailment between the document and summary units, such as sentences or dependency arcs. QGQA approaches \cite{durmus-etal-2020-feqa, wang-etal-2020-asking,scialom-etal-2021-questeval,fabbri-etal-2022-qafacteval} generates questions from one source using a question generation model and then in turn uses a question answering model to answer the generated questions given the other source. Additionally, counterfactual estimation \cite{xie-etal-2021-factual-consistency} and embedding-based metrics \cite{bert-score} have been explored. While significant progress has been made, the proposed metrics rely only on the document to detect hallucinations and ignore the other modalities. We thus propose \METRIC{} that addresses the missing modalities while maintaining similar or higher correlations with human judgment of factuality for the document and mulitmodal summarization task.
Meta-evaluations have also been proposed to evaluate such metrics for text summarization that differ in sizes and datasets \cite{fabbri-etal-2021-summeval,maynez-etal-2020-faithfulness,wang-etal-2020-asking,kryscinski-etal-2020-evaluating,pagnoni-etal-2021-understanding}. Our \METAEVAL{} is a similar effort but is the first meta-evaluation proposed for the multi-modal summarization task.

\section{Conclusion}
In this work, we present \METRIC{}, an automatic metric for evaluating factuality for multimodal abstractive summarization. Through meta-evaluation with \METAEVAL{} and additional factuality benchmarks, we show \METRIC{} and its modules correlate well with the human judgment of factuality with respect to the document, image and combined. \METRIC{} is robust across the different image and text domains and achieves competitive correlation in the zero-shot setting with more complex metrics. We hope this work provides a meta-evaluation for evaluating future multimodal factuality metrics with \METAEVAL{}, a strong baseline metric \METRIC{} to compare against, and two methods to improve the factuality of multimodal abstractive summarization models.

\section{Limitations}
We limit our work to the task that only contains the vision modality through images and the text modality. However, we note that multimodal summarization also contains video and audio, which we leave for future works.
Furthermore, similar to all pre-training models, CLIPScore and BERTScore are also known for reflecting biases
of the pre-training data \cite{hessel-etal-2021-clipscore, evaluating_clip}, leading to some incorrect predictions.
Our work is also focused for datasets in English. \citet{ladhak-etal-2020-wikilingua} proposed a multi-lingual WikiHow by aligning the steps from various languages with the image, and thus our work could be extended to include other languages by including the images to that dataset.

\section*{Acknowledgment}
We thank the reviewers for their helpful comments. We also thank Shiyue Zhang for useful discussions and comments on the paper.
This work was supported by NSF-CAREER Award 1846185, ARO Award
W911NF2110220, and NSF-AI Engage Institute DRL-211263. The
views, opinions, and/or findings contained in this article are those of the authors and not of the
funding agency.

\bibliographystyle{acl_natbib}
\bibliography{anthology,custom}

\begin{thebibliography}{55}
\expandafter\ifx\csname natexlab\endcsname\relax\def\natexlab#1{#1}\fi

\bibitem[{Agarwal et~al.(2021)Agarwal, Krueger, Clark, Radford, Kim, and
  Brundage}]{evaluating_clip}
Sandhini Agarwal, Gretchen Krueger, Jack Clark, Alec Radford, Jong~Wook Kim,
  and Miles Brundage. 2021.
\newblock \href {https://doi.org/10.48550/ARXIV.2108.02818} {Evaluating clip:
  Towards characterization of broader capabilities and downstream
  implications}.

\bibitem[{Durmus et~al.(2020)Durmus, He, and Diab}]{durmus-etal-2020-feqa}
Esin Durmus, He~He, and Mona Diab. 2020.
\newblock \href {https://doi.org/10.18653/v1/2020.acl-main.454} {{FEQA}: A
  question answering evaluation framework for faithfulness assessment in
  abstractive summarization}.
\newblock In \emph{Proceedings of the 58th Annual Meeting of the Association
  for Computational Linguistics}, pages 5055--5070, Online. Association for
  Computational Linguistics.

\bibitem[{Efron and Tibshirani(1993)}]{EfroTibs93}
Bradley Efron and Robert~J. Tibshirani. 1993.
\newblock \emph{An Introduction to the Bootstrap}.
\newblock Number~57 in Monographs on Statistics and Applied Probability.
  Chapman \& Hall/CRC, Boca Raton, Florida, USA.

\bibitem[{Fabbri et~al.(2022)Fabbri, Wu, Liu, and
  Xiong}]{fabbri-etal-2022-qafacteval}
Alexander Fabbri, Chien-Sheng Wu, Wenhao Liu, and Caiming Xiong. 2022.
\newblock \href {https://doi.org/10.18653/v1/2022.naacl-main.187}
  {{QAF}act{E}val: Improved {QA}-based factual consistency evaluation for
  summarization}.
\newblock In \emph{Proceedings of the 2022 Conference of the North American
  Chapter of the Association for Computational Linguistics: Human Language
  Technologies}, pages 2587--2601, Seattle, United States. Association for
  Computational Linguistics.

\bibitem[{Fabbri et~al.(2021)Fabbri, Kry{\'s}ci{\'n}ski, McCann, Xiong, Socher,
  and Radev}]{fabbri-etal-2021-summeval}
Alexander~R. Fabbri, Wojciech Kry{\'s}ci{\'n}ski, Bryan McCann, Caiming Xiong,
  Richard Socher, and Dragomir Radev. 2021.
\newblock \href {https://doi.org/10.1162/tacl_a_00373} {{S}umm{E}val:
  Re-evaluating summarization evaluation}.
\newblock \emph{Transactions of the Association for Computational Linguistics},
  9:391--409.

\bibitem[{Falke et~al.(2019)Falke, Ribeiro, Utama, Dagan, and
  Gurevych}]{falke-etal-2019-ranking}
Tobias Falke, Leonardo F.~R. Ribeiro, Prasetya~Ajie Utama, Ido Dagan, and Iryna
  Gurevych. 2019.
\newblock \href {https://doi.org/10.18653/v1/P19-1213} {Ranking generated
  summaries by correctness: An interesting but challenging application for
  natural language inference}.
\newblock In \emph{Proceedings of the 57th Annual Meeting of the Association
  for Computational Linguistics}, pages 2214--2220, Florence, Italy.
  Association for Computational Linguistics.

\bibitem[{Fleiss(1971)}]{Fleiss1971MeasuringNS}
Joseph~L. Fleiss. 1971.
\newblock Measuring nominal scale agreement among many raters.
\newblock \emph{Psychological Bulletin}, 76:378--382.

\bibitem[{Goyal and Durrett(2021)}]{goyal-durrett-2021-annotating}
Tanya Goyal and Greg Durrett. 2021.
\newblock \href {https://doi.org/10.18653/v1/2021.naacl-main.114} {Annotating
  and modeling fine-grained factuality in summarization}.
\newblock In \emph{Proceedings of the 2021 Conference of the North American
  Chapter of the Association for Computational Linguistics: Human Language
  Technologies}, pages 1449--1462, Online. Association for Computational
  Linguistics.

\bibitem[{Graff et~al.(2003)Graff, Kong, Chen, and Maeda}]{graff2003english}
David Graff, Junbo Kong, Ke~Chen, and Kazuaki Maeda. 2003.
\newblock English gigaword.
\newblock \emph{Linguistic Data Consortium, Philadelphia}, 4(1):34.

\bibitem[{Hermann et~al.(2015)Hermann, Kocisky, Grefenstette, Espeholt, Kay,
  Suleyman, and Blunsom}]{NIPS2015_afdec700}
Karl~Moritz Hermann, Tomas Kocisky, Edward Grefenstette, Lasse Espeholt, Will
  Kay, Mustafa Suleyman, and Phil Blunsom. 2015.
\newblock \href
  {https://proceedings.neurips.cc/paper/2015/file/afdec7005cc9f14302cd0474fd0f3c96-Paper.pdf}
  {Teaching machines to read and comprehend}.
\newblock In \emph{Advances in Neural Information Processing Systems},
  volume~28. Curran Associates, Inc.

\bibitem[{Hessel et~al.(2021)Hessel, Holtzman, Forbes, Le~Bras, and
  Choi}]{hessel-etal-2021-clipscore}
Jack Hessel, Ari Holtzman, Maxwell Forbes, Ronan Le~Bras, and Yejin Choi. 2021.
\newblock \href {https://doi.org/10.18653/v1/2021.emnlp-main.595} {{CLIPS}core:
  A reference-free evaluation metric for image captioning}.
\newblock In \emph{Proceedings of the 2021 Conference on Empirical Methods in
  Natural Language Processing}, pages 7514--7528, Online and Punta Cana,
  Dominican Republic. Association for Computational Linguistics.

\bibitem[{Hochreiter and Schmidhuber(1997)}]{10.1162/neco.1997.9.8.1735}
Sepp Hochreiter and J\"{u}rgen Schmidhuber. 1997.
\newblock \href {https://doi.org/10.1162/neco.1997.9.8.1735} {Long short-term
  memory}.
\newblock \emph{Neural Comput.}, 9(8):1735–1780.

\bibitem[{Hu et~al.(2019)Hu, Misra, and van~der Maaten}]{hexiang2018bison}
Hexiang Hu, Ishan Misra, and Laurens van~der Maaten. 2019.
\newblock {Binary Image Selection (BISON): Interpretable Evaluation of Visual
  Grounding}.
\newblock \emph{arXiv preprint arXiv:1901.06595}.

\bibitem[{Johnson et~al.(2019)Johnson, Douze, and
  J{\'e}gou}]{johnson2019billion}
Jeff Johnson, Matthijs Douze, and Herv{\'e} J{\'e}gou. 2019.
\newblock Billion-scale similarity search with {GPUs}.
\newblock \emph{IEEE Transactions on Big Data}, 7(3):535--547.

\bibitem[{Koupaee and Wang(2018)}]{wikihow}
Mahnaz Koupaee and William~Yang Wang. 2018.
\newblock \href {https://doi.org/10.48550/ARXIV.1810.09305} {Wikihow: A large
  scale text summarization dataset}.

\bibitem[{Kryscinski et~al.(2020)Kryscinski, McCann, Xiong, and
  Socher}]{kryscinski-etal-2020-evaluating}
Wojciech Kryscinski, Bryan McCann, Caiming Xiong, and Richard Socher. 2020.
\newblock \href {https://doi.org/10.18653/v1/2020.emnlp-main.750} {Evaluating
  the factual consistency of abstractive text summarization}.
\newblock In \emph{Proceedings of the 2020 Conference on Empirical Methods in
  Natural Language Processing (EMNLP)}, pages 9332--9346, Online. Association
  for Computational Linguistics.

\bibitem[{Ladhak et~al.(2020)Ladhak, Durmus, Cardie, and
  McKeown}]{ladhak-etal-2020-wikilingua}
Faisal Ladhak, Esin Durmus, Claire Cardie, and Kathleen McKeown. 2020.
\newblock \href {https://doi.org/10.18653/v1/2020.findings-emnlp.360}
  {{W}iki{L}ingua: A new benchmark dataset for cross-lingual abstractive
  summarization}.
\newblock In \emph{Findings of the Association for Computational Linguistics:
  EMNLP 2020}, pages 4034--4048, Online. Association for Computational
  Linguistics.

\bibitem[{Landis and Koch(1977)}]{Landis77}
J.~Richard Landis and Gary~G. Koch. 1977.
\newblock The measurement of observer agreement for categorical data.
\newblock \emph{Biometrics}, 33(1).

\bibitem[{Lee et~al.(2018)Lee, Chen, Hua, Hu, and He}]{lee2018stacked}
Kuang-Huei Lee, Xi~Chen, Gang Hua, Houdong Hu, and Xiaodong He. 2018.
\newblock Stacked cross attention for image-text matching.
\newblock \emph{arXiv preprint arXiv:1803.08024}.

\bibitem[{Lewis et~al.(2020)Lewis, Liu, Goyal, Ghazvininejad, Mohamed, Levy,
  Stoyanov, and Zettlemoyer}]{lewis-etal-2020-bart}
Mike Lewis, Yinhan Liu, Naman Goyal, Marjan Ghazvininejad, Abdelrahman Mohamed,
  Omer Levy, Veselin Stoyanov, and Luke Zettlemoyer. 2020.
\newblock \href {https://doi.org/10.18653/v1/2020.acl-main.703} {{BART}:
  Denoising sequence-to-sequence pre-training for natural language generation,
  translation, and comprehension}.
\newblock In \emph{Proceedings of the 58th Annual Meeting of the Association
  for Computational Linguistics}, pages 7871--7880, Online. Association for
  Computational Linguistics.

\bibitem[{Li et~al.(2020{\natexlab{a}})Li, Yuan, Xu, Wu, He, and
  Zhou}]{Li_Yuan_Xu_Wu_He_Zhou_2020}
Haoran Li, Peng Yuan, Song Xu, Youzheng Wu, Xiaodong He, and Bowen Zhou.
  2020{\natexlab{a}}.
\newblock \href {https://doi.org/10.1609/aaai.v34i05.6332} {Aspect-aware
  multimodal summarization for chinese e-commerce products}.
\newblock \emph{Proceedings of the AAAI Conference on Artificial Intelligence},
  34(05):8188--8195.

\bibitem[{Li et~al.(2018)Li, Zhu, Liu, Zhang, and Zong}]{ijcai2018-577}
Haoran Li, Junnan Zhu, Tianshang Liu, Jiajun Zhang, and Chengqing Zong. 2018.
\newblock \href {https://doi.org/10.24963/ijcai.2018/577} {Multi-modal sentence
  summarization with modality attention and image filtering}.
\newblock In \emph{Proceedings of the Twenty-Seventh International Joint
  Conference on Artificial Intelligence, {IJCAI-18}}, pages 4152--4158.
  International Joint Conferences on Artificial Intelligence Organization.

\bibitem[{Li et~al.(2020{\natexlab{b}})Li, Zhu, Zhang, He, and
  Zong}]{li-etal-2020-multimodal}
Haoran Li, Junnan Zhu, Jiajun Zhang, Xiaodong He, and Chengqing Zong.
  2020{\natexlab{b}}.
\newblock \href {https://doi.org/10.18653/v1/2020.coling-main.496} {Multimodal
  sentence summarization via multimodal selective encoding}.
\newblock In \emph{Proceedings of the 28th International Conference on
  Computational Linguistics}, pages 5655--5667, Barcelona, Spain (Online).
  International Committee on Computational Linguistics.

\bibitem[{Li et~al.(2020{\natexlab{c}})Li, Chen, Gao, Chan, Zhao, and
  Yan}]{li-etal-2020-vmsmo}
Mingzhe Li, Xiuying Chen, Shen Gao, Zhangming Chan, Dongyan Zhao, and Rui Yan.
  2020{\natexlab{c}}.
\newblock \href {https://doi.org/10.18653/v1/2020.emnlp-main.752} {{VMSMO}:
  Learning to generate multimodal summary for video-based news articles}.
\newblock In \emph{Proceedings of the 2020 Conference on Empirical Methods in
  Natural Language Processing (EMNLP)}, pages 9360--9369, Online. Association
  for Computational Linguistics.

\bibitem[{Li et~al.(2019)Li, Lei, Qin, and Wang}]{li-etal-2019-deep}
Siyao Li, Deren Lei, Pengda Qin, and William~Yang Wang. 2019.
\newblock \href {https://doi.org/10.18653/v1/D19-1623} {Deep reinforcement
  learning with distributional semantic rewards for abstractive summarization}.
\newblock In \emph{Proceedings of the 2019 Conference on Empirical Methods in
  Natural Language Processing and the 9th International Joint Conference on
  Natural Language Processing (EMNLP-IJCNLP)}, pages 6038--6044, Hong Kong,
  China. Association for Computational Linguistics.

\bibitem[{Liu et~al.(2019)Liu, Ott, Goyal, Du, Joshi, Chen, Levy, Lewis,
  Zettlemoyer, and Stoyanov}]{roberta}
Yinhan Liu, Myle Ott, Naman Goyal, Jingfei Du, Mandar Joshi, Danqi Chen, Omer
  Levy, Mike Lewis, Luke Zettlemoyer, and Veselin Stoyanov. 2019.
\newblock \href {https://doi.org/10.48550/ARXIV.1907.11692} {Roberta: A
  robustly optimized bert pretraining approach}.

\bibitem[{Loshchilov and Hutter(2019)}]{loshchilov2018decoupled}
Ilya Loshchilov and Frank Hutter. 2019.
\newblock \href {https://openreview.net/forum?id=Bkg6RiCqY7} {Decoupled weight
  decay regularization}.
\newblock In \emph{International Conference on Learning Representations}.

\bibitem[{Maynez et~al.(2020)Maynez, Narayan, Bohnet, and
  McDonald}]{maynez-etal-2020-faithfulness}
Joshua Maynez, Shashi Narayan, Bernd Bohnet, and Ryan McDonald. 2020.
\newblock \href {https://doi.org/10.18653/v1/2020.acl-main.173} {On
  faithfulness and factuality in abstractive summarization}.
\newblock In \emph{Proceedings of the 58th Annual Meeting of the Association
  for Computational Linguistics}, pages 1906--1919, Online. Association for
  Computational Linguistics.

\bibitem[{Narayan et~al.(2018)Narayan, Cohen, and
  Lapata}]{narayan-etal-2018-dont}
Shashi Narayan, Shay~B. Cohen, and Mirella Lapata. 2018.
\newblock \href {https://doi.org/10.18653/v1/D18-1206} {Don{'}t give me the
  details, just the summary! topic-aware convolutional neural networks for
  extreme summarization}.
\newblock In \emph{Proceedings of the 2018 Conference on Empirical Methods in
  Natural Language Processing}, pages 1797--1807, Brussels, Belgium.
  Association for Computational Linguistics.

\bibitem[{Oord et~al.(2018)Oord, Li, and Vinyals}]{infonce}
Aaron van~den Oord, Yazhe Li, and Oriol Vinyals. 2018.
\newblock \href {https://doi.org/10.48550/ARXIV.1807.03748} {Representation
  learning with contrastive predictive coding}.

\bibitem[{Pagnoni et~al.(2021)Pagnoni, Balachandran, and
  Tsvetkov}]{pagnoni-etal-2021-understanding}
Artidoro Pagnoni, Vidhisha Balachandran, and Yulia Tsvetkov. 2021.
\newblock \href {https://doi.org/10.18653/v1/2021.naacl-main.383}
  {Understanding factuality in abstractive summarization with {FRANK}: A
  benchmark for factuality metrics}.
\newblock In \emph{Proceedings of the 2021 Conference of the North American
  Chapter of the Association for Computational Linguistics: Human Language
  Technologies}, pages 4812--4829, Online. Association for Computational
  Linguistics.

\bibitem[{Palaskar et~al.(2019)Palaskar, Libovick{\'y}, Gella, and
  Metze}]{palaskar-etal-2019-multimodal}
Shruti Palaskar, Jind{\v{r}}ich Libovick{\'y}, Spandana Gella, and Florian
  Metze. 2019.
\newblock \href {https://doi.org/10.18653/v1/P19-1659} {Multimodal abstractive
  summarization for how2 videos}.
\newblock In \emph{Proceedings of the 57th Annual Meeting of the Association
  for Computational Linguistics}, pages 6587--6596, Florence, Italy.
  Association for Computational Linguistics.

\bibitem[{Parnell et~al.(2021)Parnell, Jauregi~Unanue, and
  Piccardi}]{parnell-etal-2021-rewardsofsum}
Jacob Parnell, Inigo Jauregi~Unanue, and Massimo Piccardi. 2021.
\newblock \href {https://doi.org/10.18653/v1/2021.spnlp-1.1}
  {{R}ewards{O}f{S}um: Exploring reinforcement learning rewards for
  summarisation}.
\newblock In \emph{Proceedings of the 5th Workshop on Structured Prediction for
  NLP (SPNLP 2021)}, pages 1--11, Online. Association for Computational
  Linguistics.

\bibitem[{Pasunuru and Bansal(2018)}]{pasunuru-bansal-2018-multi}
Ramakanth Pasunuru and Mohit Bansal. 2018.
\newblock \href {https://doi.org/10.18653/v1/N18-2102} {Multi-reward reinforced
  summarization with saliency and entailment}.
\newblock In \emph{Proceedings of the 2018 Conference of the North {A}merican
  Chapter of the Association for Computational Linguistics: Human Language
  Technologies, Volume 2 (Short Papers)}, pages 646--653, New Orleans,
  Louisiana. Association for Computational Linguistics.

\bibitem[{Radford et~al.(2021)Radford, Kim, Hallacy, Ramesh, Goh, Agarwal,
  Sastry, Askell, Mishkin, Clark, Krueger, and Sutskever}]{clip}
Alec Radford, Jong~Wook Kim, Chris Hallacy, Aditya Ramesh, Gabriel Goh,
  Sandhini Agarwal, Girish Sastry, Amanda Askell, Pamela Mishkin, Jack Clark,
  Gretchen Krueger, and Ilya Sutskever. 2021.
\newblock \href {http://arxiv.org/abs/2103.00020} {Learning transferable visual
  models from natural language supervision}.
\newblock \emph{CoRR}, abs/2103.00020.

\bibitem[{Raffel et~al.(2020)Raffel, Shazeer, Roberts, Lee, Narang, Matena,
  Zhou, Li, and Liu}]{JMLR:v21:20-074}
Colin Raffel, Noam Shazeer, Adam Roberts, Katherine Lee, Sharan Narang, Michael
  Matena, Yanqi Zhou, Wei Li, and Peter~J. Liu. 2020.
\newblock \href {http://jmlr.org/papers/v21/20-074.html} {Exploring the limits
  of transfer learning with a unified text-to-text transformer}.
\newblock \emph{Journal of Machine Learning Research}, 21(140):1--67.

\bibitem[{Rennie et~al.(2017)Rennie, Marcheret, Mroueh, Ross, and
  Goel}]{8099614}
Steven~J. Rennie, Etienne Marcheret, Youssef Mroueh, Jerret Ross, and Vaibhava
  Goel. 2017.
\newblock \href {https://doi.org/10.1109/CVPR.2017.131} {Self-critical sequence
  training for image captioning}.
\newblock In \emph{2017 IEEE Conference on Computer Vision and Pattern
  Recognition (CVPR)}, pages 1179--1195.

\bibitem[{Rush et~al.(2015)Rush, Chopra, and Weston}]{rush-etal-2015-neural}
Alexander~M. Rush, Sumit Chopra, and Jason Weston. 2015.
\newblock \href {https://doi.org/10.18653/v1/D15-1044} {A neural attention
  model for abstractive sentence summarization}.
\newblock In \emph{Proceedings of the 2015 Conference on Empirical Methods in
  Natural Language Processing}, pages 379--389, Lisbon, Portugal. Association
  for Computational Linguistics.

\bibitem[{Scialom et~al.(2021)Scialom, Dray, Lamprier, Piwowarski, Staiano,
  Wang, and Gallinari}]{scialom-etal-2021-questeval}
Thomas Scialom, Paul-Alexis Dray, Sylvain Lamprier, Benjamin Piwowarski, Jacopo
  Staiano, Alex Wang, and Patrick Gallinari. 2021.
\newblock \href {https://doi.org/10.18653/v1/2021.emnlp-main.529}
  {{Q}uest{E}val: Summarization asks for fact-based evaluation}.
\newblock In \emph{Proceedings of the 2021 Conference on Empirical Methods in
  Natural Language Processing}, pages 6594--6604, Online and Punta Cana,
  Dominican Republic. Association for Computational Linguistics.

\bibitem[{Shekhar et~al.(2017)Shekhar, Pezzelle, Klimovich, Herbelot, Nabi,
  Sangineto, and Bernardi}]{shekhar-etal-2017-foil}
Ravi Shekhar, Sandro Pezzelle, Yauhen Klimovich, Aur{\'e}lie Herbelot, Moin
  Nabi, Enver Sangineto, and Raffaella Bernardi. 2017.
\newblock \href {https://doi.org/10.18653/v1/P17-1024} {{FOIL} it! find one
  mismatch between image and language caption}.
\newblock In \emph{Proceedings of the 55th Annual Meeting of the Association
  for Computational Linguistics (Volume 1: Long Papers)}, pages 255--265,
  Vancouver, Canada. Association for Computational Linguistics.

\bibitem[{Simonyan and Zisserman(2015)}]{Simonyan15}
Karen Simonyan and Andrew Zisserman. 2015.
\newblock Very deep convolutional networks for large-scale image recognition.
\newblock In \emph{International Conference on Learning Representations}.

\bibitem[{Sung et~al.(2022)Sung, Cho, and Bansal}]{sung2022vladapter}
Yi-Lin Sung, Jaemin Cho, and Mohit Bansal. 2022.
\newblock Vl-adapter: Parameter-efficient transfer learning for
  vision-and-language tasks.
\newblock In \emph{CVPR}.

\bibitem[{Vassileios~Balntas and Mikolajczyk(2016)}]{BMVC2016_119}
Daniel~Ponsa Vassileios~Balntas, Edgar~Riba and Krystian Mikolajczyk. 2016.
\newblock \href {https://doi.org/10.5244/C.30.119} {Learning local feature
  descriptors with triplets and shallow convolutional neural networks}.
\newblock In \emph{Proceedings of the British Machine Vision Conference
  (BMVC)}, pages 119.1--119.11. BMVA Press.

\bibitem[{Wang et~al.(2020)Wang, Cho, and Lewis}]{wang-etal-2020-asking}
Alex Wang, Kyunghyun Cho, and Mike Lewis. 2020.
\newblock \href {https://doi.org/10.18653/v1/2020.acl-main.450} {Asking and
  answering questions to evaluate the factual consistency of summaries}.
\newblock In \emph{Proceedings of the 58th Annual Meeting of the Association
  for Computational Linguistics}, pages 5008--5020, Online. Association for
  Computational Linguistics.

\bibitem[{Williams et~al.(2018)Williams, Nangia, and
  Bowman}]{williams-etal-2018-broad}
Adina Williams, Nikita Nangia, and Samuel Bowman. 2018.
\newblock \href {https://doi.org/10.18653/v1/N18-1101} {A broad-coverage
  challenge corpus for sentence understanding through inference}.
\newblock In \emph{Proceedings of the 2018 Conference of the North {A}merican
  Chapter of the Association for Computational Linguistics: Human Language
  Technologies, Volume 1 (Long Papers)}, pages 1112--1122, New Orleans,
  Louisiana. Association for Computational Linguistics.

\bibitem[{Williams(1992)}]{10.1007/BF00992696}
Ronald~J. Williams. 1992.
\newblock \href {https://doi.org/10.1007/BF00992696} {Simple statistical
  gradient-following algorithms for connectionist reinforcement learning}.
\newblock \emph{Mach. Learn.}, 8(3–4):229–256.

\bibitem[{Wolf et~al.(2020)Wolf, Debut, Sanh, Chaumond, Delangue, Moi, Cistac,
  Rault, Louf, Funtowicz, Davison, Shleifer, von Platen, Ma, Jernite, Plu, Xu,
  Le~Scao, Gugger, Drame, Lhoest, and Rush}]{wolf-etal-2020-transformers}
Thomas Wolf, Lysandre Debut, Victor Sanh, Julien Chaumond, Clement Delangue,
  Anthony Moi, Pierric Cistac, Tim Rault, Remi Louf, Morgan Funtowicz, Joe
  Davison, Sam Shleifer, Patrick von Platen, Clara Ma, Yacine Jernite, Julien
  Plu, Canwen Xu, Teven Le~Scao, Sylvain Gugger, Mariama Drame, Quentin Lhoest,
  and Alexander Rush. 2020.
\newblock \href {https://doi.org/10.18653/v1/2020.emnlp-demos.6} {Transformers:
  State-of-the-art natural language processing}.
\newblock In \emph{Proceedings of the 2020 Conference on Empirical Methods in
  Natural Language Processing: System Demonstrations}, pages 38--45, Online.
  Association for Computational Linguistics.

\bibitem[{Xiao and Wang(2021)}]{xiao-wang-2021-hallucination}
Yijun Xiao and William~Yang Wang. 2021.
\newblock \href {https://doi.org/10.18653/v1/2021.eacl-main.236} {On
  hallucination and predictive uncertainty in conditional language generation}.
\newblock In \emph{Proceedings of the 16th Conference of the European Chapter
  of the Association for Computational Linguistics: Main Volume}, pages
  2734--2744, Online. Association for Computational Linguistics.

\bibitem[{Xie et~al.(2021)Xie, Sun, Deng, Li, and
  Ding}]{xie-etal-2021-factual-consistency}
Yuexiang Xie, Fei Sun, Yang Deng, Yaliang Li, and Bolin Ding. 2021.
\newblock \href {https://doi.org/10.18653/v1/2021.findings-emnlp.10} {Factual
  consistency evaluation for text summarization via counterfactual estimation}.
\newblock In \emph{Findings of the Association for Computational Linguistics:
  EMNLP 2021}, pages 100--110, Punta Cana, Dominican Republic. Association for
  Computational Linguistics.

\bibitem[{Yang et~al.(2021)Yang, Panagopoulou, Lyu, Zhang, Yatskar, and
  Callison-Burch}]{yang-etal-2021-visual}
Yue Yang, Artemis Panagopoulou, Qing Lyu, Li~Zhang, Mark Yatskar, and Chris
  Callison-Burch. 2021.
\newblock \href {https://doi.org/10.18653/v1/2021.emnlp-main.165} {Visual
  goal-step inference using wiki{H}ow}.
\newblock In \emph{Proceedings of the 2021 Conference on Empirical Methods in
  Natural Language Processing}, pages 2167--2179, Online and Punta Cana,
  Dominican Republic. Association for Computational Linguistics.

\bibitem[{Zhang et~al.(2020)Zhang, Zhao, Saleh, and Liu}]{pmlr-v119-zhang20ae}
Jingqing Zhang, Yao Zhao, Mohammad Saleh, and Peter Liu. 2020.
\newblock \href {https://proceedings.mlr.press/v119/zhang20ae.html} {{PEGASUS}:
  Pre-training with extracted gap-sentences for abstractive summarization}.
\newblock In \emph{Proceedings of the 37th International Conference on Machine
  Learning}, volume 119 of \emph{Proceedings of Machine Learning Research},
  pages 11328--11339. PMLR.

\bibitem[{Zhang* et~al.(2020)Zhang*, Kishore*, Wu*, Weinberger, and
  Artzi}]{bert-score}
Tianyi Zhang*, Varsha Kishore*, Felix Wu*, Kilian~Q. Weinberger, and Yoav
  Artzi. 2020.
\newblock \href {https://openreview.net/forum?id=SkeHuCVFDr} {Bertscore:
  Evaluating text generation with bert}.
\newblock In \emph{International Conference on Learning Representations}.

\bibitem[{Zhang et~al.(2022)Zhang, Meng, Wang, Jiang, Liu, and Yang}]{unims}
Zhengkun Zhang, Xiaojun Meng, Yasheng Wang, Xin Jiang, Qun Liu, and Zhenglu
  Yang. 2022.
\newblock Unims: A unified framework for multimodal summarization with
  knowledge distillation.
\newblock \emph{Proceedings of the AAAI Conference on Artificial Intelligence}.

\bibitem[{Zhu et~al.(2018)Zhu, Li, Liu, Zhou, Zhang, and
  Zong}]{zhu-etal-2018-msmo}
Junnan Zhu, Haoran Li, Tianshang Liu, Yu~Zhou, Jiajun Zhang, and Chengqing
  Zong. 2018.
\newblock \href {https://doi.org/10.18653/v1/D18-1448} {{MSMO}: Multimodal
  summarization with multimodal output}.
\newblock In \emph{Proceedings of the 2018 Conference on Empirical Methods in
  Natural Language Processing}, pages 4154--4164, Brussels, Belgium.
  Association for Computational Linguistics.

\bibitem[{Zhu et~al.(2020)Zhu, Zhou, Zhang, Li, Zong, and
  Li}]{Zhu_Zhou_Zhang_Li_Zong_Li_2020}
Junnan Zhu, Yu~Zhou, Jiajun Zhang, Haoran Li, Chengqing Zong, and Changliang
  Li. 2020.
\newblock \href {https://doi.org/10.1609/aaai.v34i05.6525} {Multimodal
  summarization with guidance of multimodal reference}.
\newblock \emph{Proceedings of the AAAI Conference on Artificial Intelligence},
  34(05):9749--9756.

\end{thebibliography}

\appendix

\section{Meta-Evaluation Details}

\begin{table*}[!t]
    \centering
    \small
    \begin{tabular}{c c c c c c c c}
    \toprule
    Model & Optimizer & Learning rate & Label Smoothing & Num steps & Batch size \\
    \midrule
    T5 & AdamW & 5e-5 & 0.1 & 15,000 & 256\\
    PEGASUS & AdaFactor & 8e-5 & 0.1 & 15,000 & 256 \\
    CLIP-BART & AdamW & 5e-5 & 0.1 & 15,000 & 256 \\
    MOF & Adam & 1e-3 & 0.0 & 50,000 & 512 \\
    \bottomrule
    \end{tabular}
    \caption{Hyper-parameters of the models trained on the multimodal WikiHow summarization task.}
    \label{tab:human_annotation_models}
\end{table*}

\subsection{Metrics Details} \label{sec:metrics_details}

We describe the metrics we use for computing correlations. We use the official scripts from the respective repository.

\paragraph{FactCC.} FactCC \cite{kryscinski-etal-2020-evaluating} is an entailment-based metric that uses BERT to output a binary prediction of factuality given the concatenation of the document and a summary sentence as input. The final score is the average factuality score of all summary sentences.

\paragraph{DAE.} DAE \cite{goyal-durrett-2021-annotating} is an entailment-based metric that evaluates factuality on dependency arc level instead of on sentence level. We report the sentence-level error. The sentence is considered to contain an error if any of its arcs are predicted to be non-factual, and the final score is the average of all sentence predictions. 0 indicates a sentence contains no error, and 1 indicates the sentence contains an error.

\paragraph{QuestEval.} A QGQA metric, \citet{scialom-etal-2021-questeval} generates questions using a question generation model from both the document and the summary. Then, a question-answering model answers the question generated using the document with the summary, and answers the question generated using the summary with the document. The final score is the harmonic mean of the accuracy of the predicted answers to the true answers from the question generation model.

\paragraph{CLIPText-S.} CLIPScore provides a variant of the metric that takes in references for the image captioning, and calculates the cosine similarity between the text embeddings $T$ and that of the references $R$. The final score is calculated by taking the average over the maximum reference cosine similarity:
$$ \text{CLIPText-S}(T,R) = \frac{1}{|T|} \sum_{i=1}^{|T|} \max_{r \in R} \text{cossim}(v_i, r) $$

\paragraph{BERTScore/BERT-S.} The original BERTScore uses \texttt{roberta-large} by default. For BERT-S, we use \texttt{roberta-large-mnli} up to the 11th layer after tuning on FRANK's validation set.

\paragraph{Triplet Network.} This network maps image and summary embeddings into the same space and minimize the distance between the positive pair and maximize that between the pair of image and negative summaries with the Triplet loss \cite{BMVC2016_119}. Specifically, a triplet Network takes in a triplet $(V,S_{pos}, S_{neg})$, the representation of an image $V$, and that of a positive summary and negative image. We then map the representation to the same space and normalize the embeddings. We then use the triplet loss with a margin of 0.2. To generate the negative summaries, we use the similarity split of VGSI and take the summaries for the three negative choices.
We use the CLIP \texttt{RN50x64} visual backbone to generate the visual representation and use BERT to generate the summary representation. We modify the example training code provided by Transformers, and train for 10 epochs with a learning rate of 5e-5. We use the other default settings.

\paragraph{MMAE.} MMAE \cite{zhu-etal-2018-msmo} is initially developed for evaluating the summary quality on MSMO, which we adapt to our task. The metric consists of three submodules: image precision (IP), IM\textsubscript{MAX}, and ROUGE-L. For  \METAEVAL{}, since we only have a  single image, IP is just 1. IM\textsubscript{MAX} is trained on the multimodal WikiHow dataset, where the negative image-summary pair is from the same batch. We use the same hyper-parameters of the original MMAE metric. To combine the three scores, we use the MLP variant tuned on the validation set of \METAEVAL{}.

\paragraph{Combined Metrics.} We tune the combined metrics on the validation set of \METAEVAL{}. We use $\alpha=0.45$ for CLIP-S + DAE, $\alpha=0.10$ for Triplet Net + BERT-S, and $0.25$ for \METRIC{}.

\subsection{Model Details} \label{sec:human_annotation_models}
We train the models on the proposed multimodal WikiHow dataset. The hyper-parameters are shown in \autoref{tab:human_annotation_models}. The pre-trained models and the training scripts  for the transformer-based models are taken from HuggingFace's transformers library \cite{wolf-etal-2020-transformers}. We set the maximum input length to 256 and output length to 32 for all models.

\paragraph{T5.} T5 \cite{JMLR:v21:20-074} is an encoder-decoder model pre-trained on a collection of tasks in a text-to-text format. We use the \texttt{t5-small} model and fine-tune as a document summarization tasks, ignoring the images. The total number of parameters is around 60 million. We use mixed precision, and training was performed on 2 NVIDIA RTX A6000 GPUs for approximately 6 hours.

\paragraph{PEGASUS.} PEGASUS \cite{pmlr-v119-zhang20ae} is another encoder-decoder model specifically designed for the abstractive summarization task by imitating the summarization setup during pre-training. We use \texttt{PEGASUS-large} checkpoint and fine-tune without the images. The total number of parameters is around 571 million. Training was performed on a single NVIDIA RTX A6000 GPU for approximately 28 hours.

\paragraph{CLIP-BART.} The architecture of CLIP-BART is described in Section~\ref{sec:downstream_applications}. The total number of parameters is around 140 million. We fine-tune the model starting from the \texttt{BART-base} checkpoint, and use the CLIP \texttt{RN50x64} visual encoder to extract image features. We use mixed precision, and the training was performed on a single NVIDIA RTX A6000 GPU for approximately 6 hours.

\paragraph{MOF.} MOF is based on \citet{zhu-etal-2018-msmo}, a multimodal summariaziton model with multimodal attention \cite{ijcai2018-577}. The model consists of a single-layer unidirectional LSTM \cite{10.1162/neco.1997.9.8.1735} with the embedding dimension of 256 and hidden dimension of 512 for the text encoder and text decoder. The multimodal attention is computed by concatenating the textual attention layer and visual attention layer over the visual features, extracted from the global \texttt{fc7} layers from VGG19 \cite{Simonyan15}. The total number of parameters is around 83M. Training was performed on a single NVIDIA RTX A6000 GPU for approximately 40 hours.

\subsection{Annotation Details} \label{sec:human_annotation_details}
\begin{figure*}[!t]
    \centering
    \includegraphics[width=\textwidth,keepaspectratio]{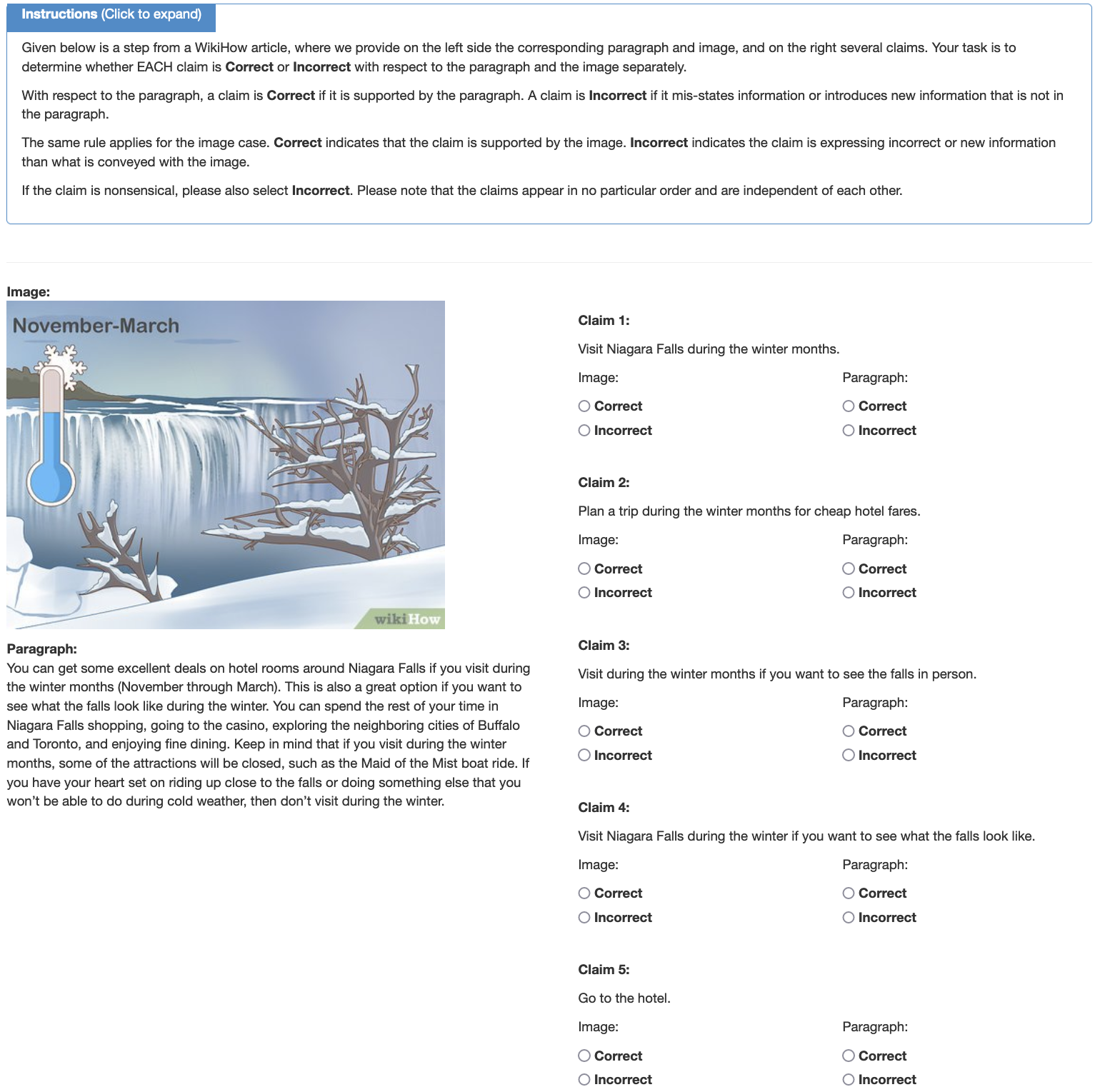}
    \caption{Screenshot of the annotation interface on AMT.}
    \label{fig:human_eval_example}
\end{figure*}

\autoref{fig:human_eval_example} shows a screenshot of the annotation task on AMT.

\begin{table}[!t]
    \centering
    \small
    \adjustbox{max width=\columnwidth}{
    \begin{tabular}{c c}
    \toprule
    Metric &  Cont. Combined \\
    \midrule
    FactCC & 0.01 \\
    DAE & 0.43 \\
    QuestEval & 0.39 \\
    CLIPText-S &  0.19 \\
    BERTScore & 0.50 \\
    BERT-S & \textbf{0.54} \\
    \midrule
    CLIP-S & 0.23 \\
    IM\textsubscript{Max} & 0.07 \\
    Triplet Net & \textbf{0.25} \\
    \midrule
    MMAE\textsubscript{Logis} & 0.27 \\
    MMAE\textsubscript{MLP} & 0.26 \\
    RefCLIP-S & 0.26 \\
    CLIP-S + DAE & 0.47 \\
    Triplet Net + BERT-S & \textbf{0.56} \\
    \METRIC{} & \textbf{0.56} \\
    \bottomrule
    \end{tabular}
    }
    \caption{Pearson correlation coefficients between automatic metrics and human judgments of factuality with respect to the continuous combined judgment.}
    \label{tab:continous_result}
\end{table}

\section{Meta-Evaluation with Continuous Labels}\label{sec:continous_judgment}
We also experiment with combining the two judgments in a continuous way, by taking the average of the two judgments so that a score of 0.5 indicates that the summary is faithful to only one modality. The combined judgment is shown in \autoref{tab:continous_result}. While the correlations are higher overall for all metrics, the trend is similar to the \autoref{tab:human_eval_result}, where \METRIC{} can match the correlations of the fine-tuned metric, Triplet Net + BERT-S, indicating the effectiveness and simplicity of our metric.

\section{Additional Factuality Benchmark Evaluations Details}\label{sec:add_benchmark_appendix}

\begin{table}[!t]
    \centering
    \small
    \begin{tabular}{c c c c}
    \toprule
    Metric & Random & Category & Similarity \\
    \midrule
    Random & 25.00 & 25.00 & 25.00 \\
    Triplet Net* & 78.48 & 74.65 & \textbf{66.07} \\
    CLIP-S ViT-B/32 & 83.05 & 75.42 & 62.86 \\
    CLIP-S & \textbf{87.79} & \textbf{81.37} & \textbf{70.94} \\
    \midrule
    Human* & 92.00 & 89.20 & 86.00 \\
    \bottomrule
    \end{tabular}
    \caption{Original Wikihow VGSI. Results with * indicates results taken from the original paper.}
    \label{tab:wikihow_vgsi_orig}
\end{table}

\subsection{WikiHowFact Details}\label{sec:wikihowfact_appendix}

\begin{table}[!t]
    \centering
    \small
    \begin{tabular}{l c}
    \toprule
    Metric & Acc \\
    \midrule
    SCAN t2i* & 85.89 \\
    Triplet Net & 63.16 \\
    CLIP-S ViT-B/32 & 83.85 \\
    CLIP-S ViT-B/16 & 85.36 \\
    CLIP-S ViT-L/14 & 85.89\\
    CLIP-S RN50 & 83.50 \\
    CLIP-S RN101 & 83.97 \\
    CLIP-S RN50x4 & 84.95 \\
    CLIP-S RN50x16 & 85.22 \\
    CLIP-S RN50x64 & \textbf{86.03} \\
    \bottomrule
    \end{tabular}
    \caption{BISON accuracy. * indicates result copied from \citet{hexiang2018bison}.}
    \label{tab:bison_full}
\end{table}

\begin{table*}[t!]
    \centering
    \small
    \adjustbox{max width=\textwidth}{
    \begin{tabular}{c | cccc | cccc | cccc }
    \toprule
     & \multicolumn{4}{c}{All data}  & \multicolumn{4}{c}{CNN/DM} & \multicolumn{4}{c}{XSum} \\
     \midrule
     \multirow{2}{*}{Metric} & \multicolumn{2}{c}{Pearson} & \multicolumn{2}{c}{Spearman} & \multicolumn{2}{c}{Pearson} & \multicolumn{2}{c}{Spearman} & \multicolumn{2}{c}{Pearson} & \multicolumn{2}{c}{Spearman} \\
     & $\rho$ & p-val & $\rho$ & p-val & $\rho$ & p-val & $\rho$ & p-val & $\rho$ & p-val & $\rho$ & p-val  \\
     \midrule
     FactCC* & 0.20 & 0.00 & \textbf{0.30} & 0.00 & 0.36 & 0.00 & 0.30 & 0.00 & 0.07 & 0.73 & 0.19 & 0.00 \\
     DAE* & 0.18 & 0.00 & 0.20 & 0.00 & 0.03 & 0.38 & 0.33 & 0.00 & \textbf{0.27} & 0.00 & \textbf{0.22} & 0.00 \\
     BERTScore* & 0.30 & 0.00 & 0.25 & 0.00 & 0.38 & 0.00 & 0.31 & 0.00 & 0.20 & 0.00 & 0.09 &	0.17 \\
     QuestEval & 0.23 & 0.00 & 0.23 & 0.00 & 0.27 & 0.00 & 0.25 & 0.00 & 0.18 & 0.00 & 0.10 & 0.00 \\
     CLIPText-S & 0.11 & 0.00 & 0.09 & 0.00 & 0.11 & 0.00 & 0.12 & 0.00 & 0.10 & 0.00 & 0.05 & 0.17 \\
     BERT-S & \textbf{0.31} & 0.00 & 0.26 & 0.00 & \textbf{0.40} & 0.00 & \textbf{0.32} & 0.00 & 0.22 & 0.00 & 0.11 & 0.00 \\
     \bottomrule
    \end{tabular}
    }
    \caption{Correlation with human judgment of factuality on FRANK. BERT-S achieves overall higher correlations than its original variant and achieves the highest Pearson correlation on all data.}
    \label{tab:frank}
\end{table*}

The three negative images are selected with three different sampling strategies, following \citet{yang-etal-2021-visual}: \textit{Random} selects the three images arbitrarily, \textit{Category} randomly selects three negative images from the same WikiHow category\footnote{\url{https://www.wikihow.com/Special:CategoryListing}}, and \textit{Similarity} selects top-3 most similar images from different articles using similarity computed using FAISS \cite{johnson2019billion}. \textit{Random} consists of 153,961 examples, \textit{similarity} consists of 153,770 examples, and \textit{category} contains 153,961 examples.

The three sampling strategies provide settings with increasing difficulty in terms of the negative summaries; random is the easiest setting and similarity is the hardest. Depending on which modality we provide as the prompt, we can further break down the task and evaluate the metric's performance with different combinations of modalities. We use the same sets of metrics described in Section~\ref{sec:human_eval} for each modality combination. FactCC and DAE produce binary labels and thus are at a disadvantage for the ranking experiment, and we thus use the probability for the factual label for FactCC and the token error for DAE. To explore how larger visual backbone can improve image-summary factuality detection, we 
compare against the original CLIP-S that uses the \texttt{ViT-B/32} backbone.

\begin{table}[!t]
    \centering
    \small
    \begin{tabular}{l c c c c}
    \toprule
    Metric & no-ref & 1-ref & 4-ref \\
    \midrule
    length* & - & 50.2 & 50.2 \\
    ROUGE-L* & - & 71.7 & 79.3 \\
    CLIPText-S ViT-B/32 & - & 87.13 & 90.58 \\
    CLIPText-S ViT-B/16 & - & 87.58 & 91.43 \\
    CLIPText-S ViT-L/14 & - & 88.52 & 92.01 \\
    CLIPText-S RN50 & - & 86.59 & 89.91 \\
    CLIPText-S RN101 & - & 86.99 & 89.50 \\
    CLIPText-S RN50x4 & - & 87.62 & 90.50 \\
    CLIPText-S RN50x16 & - & 87.79 & 91.42 \\
    CLIPText-S RN50x64 & - & 87.82 & 92.01 \\
    BERT-S & - & \textbf{87.84} & \textbf{93.59} \\
    \midrule
    Triplet Net + BERT-S & 60.47 & 84.97 & 90.74 \\
    RefCLIP-S ViT-B/32 & 86.84 & 91.70 & 92.55 \\
    RefCLIP-S ViT-B/16 & 89.00 & 91.80 & 93.37 \\
    RefCLIP-S ViT-L/14 & 89.24 & 92.58 & 93.77 \\
    RefCLIP-S RN50 & 86.15 & 91.25 & 91.83 \\
    RefCLIP-S RN101 & 86.72 & 91.84 & 93.12 \\
    RefCLIP-S RN50x4 & 87.42 & 91.83 & 93.27 \\
    RefCLIP-S RN50x16 & 88.49 & 92.09 & 93.54 \\
    RefCLIP-S RN50x64 & \textbf{90.05} & 91.95 & 93.79 \\
    CLIPBERTScore & \textbf{90.05} & \textbf{92.68} & \textbf{95.01} \\
    \bottomrule
    \end{tabular}
    \caption{FOIL accuracy. * indicates results copied from \citet{hessel-etal-2021-clipscore}.}
    \label{tab:foil_full}
\end{table}

\begin{table*}[!t]
    \centering
    \adjustbox{max width=\textwidth}{
    \begin{tabular}{c l c c c c c c}
    \toprule
    Dataset & Model & \begin{tabular}{@{}c@{}}Learning \\ rate\end{tabular} & \begin{tabular}{@{}c@{}}Label \\ smoothing\end{tabular} & \# Steps & \begin{tabular}{@{}c@{}}Batch \\ size\end{tabular} & \begin{tabular}{@{}c@{}}Max input \\ tokens\end{tabular} & \begin{tabular}{@{}c@{}}Max target \\ tokens\end{tabular} \\
    \midrule
    \multirow{2}{*}{MMSS} &  CLIP-BART & 3e-5 & 0.1 & 5,000 & 256 & 128 & 32 \\
    & CLIP-BART + RL & 3e-5 & 0.0 & 5,000 & 256 & 128 & 32 \\
    \midrule
    \multirow{3}{*}{MSMO} & CLIP-BART + ROUGE & 3e-5 & 0.1 & 20,000 & 256 & 512 & 128 \\
    & CLIP-BART + \METRIC{} & 3e-5 & 0.1 & 20,000 & 256 & 512 & 128 \\
    & CLIP-BART + \METRIC{} + RL & 3e-5 & 0.0 & 10,000 & 256 & 512 & 128 \\
    \bottomrule
    \end{tabular}
    }
    \caption{Hyper-parameters of the models trained for downstream applications.}
    \label{tab:downstream_models}
\end{table*}

\paragraph{Comparison with VGSI.} As described in Section~\ref{sec:wikihow_fact}, the difference between VGSI and WikiHowFact is what information is provided as the prompt and the choices. For VGSI, we use the step sentence, or the summary, as the prompt and ask the models to select the correct image. Since the document is not provided, we use CLIP-S to calculate the score for each summary and image pair. We show the result in \autoref{tab:wikihow_vgsi_orig}. We see the same surprising result that CLIP-S with the \texttt{ViT-B/32} backbone achieves better ranking accuracy than the Triplet Net model trained on the training split. Increasing the capacity of the CLIP-S with \texttt{RN50x64}, the ranking accuracy improves by 4 points for random, and 8 points for category and similarity, approaching the human performance, especially for the random case. Additionally, when comparing the performance of the same model for WikiHowFact and VGSI, the ranking accuracies for VGSI are much higher, indicating that WikiHowFact is more difficult.

\subsection{FOIL}\label{sec:foil_appendix}
We explore the ability of CLIP-S at differentiating hallucinating captions.
The FOIL \cite{shekhar-etal-2017-foil} dataset measures how well the metric can differentiate hallucinated captions from the correct ones. The task uses MSCOCO reference captions and adversarially swaps out noun phrases to create hallucinating summaries to create 64K test pairs.

One benefit of captioning tasks is that the captioning data contain references that can be treated as the document in our setting, and thus enable evaluation of different modality combinations similar to the multimodal summarization setting. We consider three settings, where we show no reference (evaluating only on the image-text setting), as well as providing 1 or 4 additional reference captions (excluding the true caption being evaluated).
We concatenate the references and consider them as documents. We compare \METRIC{} and its components primarily against the CLIPScore variants, including CLIPText-S and RefCLIP-S. For the image-text hallucination detection, we focus on how the different CLIP backbones affects factuality detection between image and the summary. This includes \texttt{ViT-B/32}, \texttt{ViT-B/16}, \texttt{ViT-L/14}, \texttt{RN50}, \texttt{RN101}, \texttt{RN50x4}, \texttt{RN50x16}, and \texttt{RN50x64}.

We present the results in \autoref{tab:foil_full}. We observe a clear trend that larger visual backbones improve accuracy when considering only the visual performance for the no-ref case. Interesingly, ViT-based models outperform the RN-based ones for this task.

\subsection{BISON}\label{sec:bison_appendix}
BISON \cite{hexiang2018bison} measures the ability of the metric to select the correct image from two semantically similar images, and thus assesses whether the metric is able to detect fine-grained information present in the text and image. We compare the image-summary metrics, including all CLIP-S variants, similar to FOIL (Appendix~\ref{sec:foil_appendix})

\autoref{tab:bison_full} shows the result. We observe a similar improvement in accuracy with larger visual backbones as observed with the FOIL dataset. While we similarly observe improvement as the model size grows, CLIP-S RN50x64 is the only backbone that outperforms the fully fine-tuned SOTA metric, SCAN t2i \cite{lee2018stacked}.

\begin{table}[!t]
    \centering
    \small
    \begin{tabular}{c c c c c c}
    \toprule
        Dataset & \#Train & \#Validation & \#Test & \#img \\
        \midrule
        WikiHow & 710,737 & 29,872 & 30,183 & 1 \\
        MMSS & 62,000 & 2,000 & 2,000 & 1 \\
        MSMO & 293,895 & 10,354 & 10,258 & 6.57 \\
        \bottomrule
    \end{tabular}
    \caption{Dataset Statistics}
    \label{tab:dataset_statistics}
\end{table}

\subsection{FRANK}\label{sec:frank_appendix}
We show the result in \autoref{tab:frank}. As described in Section~\ref{sec:frank}, we compare existing factuality metrics with CLIPText-S and BERT-S. We also include QuestEval, which does not correlate better than BERTScore variants. CLIPText-S does not perform well for detecting faithful summaries for summarization, as Pearson and Spearman coefficients are around 0.10 for all datasets and 0.05 for XSum Spearman. In contrast, BERT-S that uses RoBERTa \cite{roberta} model finetuned on the MNLI \cite{williams-etal-2018-broad} correlates higher than the original BERTScore on Pearson and Spearman across both datasets. It is thus useful to treat factuality as an entailment problem and use the appropriate model.

\section{Downstream Applications Details} \label{sec:rl_experiment}
For both experiments, we use the CLIP \texttt{RN50x64} visual encoder to extract the visual features and we limit the maximum number of images to 10. We train the models with transformers library. For all models, We train the models with mixed precision and AdamW \cite{loshchilov2018decoupled}. Other hyper-parameters are found in \autoref{tab:downstream_models}.

All CLIP-BART models are trained with 4 NVIDIA RTX A6000 GPUs. The training took approximately an hour for MMSS, and  approximately 19 hours for both MSMO baseline models.

For SCST training, we train the models on a single NVIDIA RTX A6000 GPU. Training took approximately 7 hours for MMSS and approximately 70 hours for MSMO. We perform a hyper-parameter search manually by evaluating the models on the validation set of the corresponding datasets and select the best-performing parameter according to BERTScore and ROUGE-2 (since these are the scores we optimize for). We first determining the $\alpha$ from \{0.90, 0.95,0.99, 0.995, 0.998, 0.999, 1.0\}, where we find 0.998 to perform the best. We then tune the weight of \METRIC{} from \{1,2,5\} and find that 2 performs the best for both datasets.

\end{document}